\documentclass{article} %
\usepackage{iclr2027_conference,times}

\usepackage{amsmath,amsfonts,bm}

\def\secref#1{section~\ref{#1}}

\def\eqref#1{equation~\ref{#1}}
\def\Eqref#1{Equation~\ref{#1}}

\def\1{\bm{1}}

\def\eps{{\epsilon}}

\DeclareMathAlphabet{\mathsfit}{\encodingdefault}{\sfdefault}{m}{sl}
\SetMathAlphabet{\mathsfit}{bold}{\encodingdefault}{\sfdefault}{bx}{n}

\newcommand{\E}{\mathbb{E}}

\newcommand{\R}{\mathbb{R}}

\DeclareMathOperator*{\argmax}{arg\,max}

\newcommand{\alphabar}{\bar{\alpha}}
\newcommand{\norm}[1]{\lVert #1 \rVert}
\newcommand{\diff}{\mathrm{d}}
\usepackage{tikz}
\newcommand*\circled[1]{\tikz[baseline=(char.base)]{%
    \node[shape=circle,fill=black,draw,inner sep=1pt,text=white] (char) {#1};}}

\usepackage{hyperref}
\usepackage{url}
\usepackage{booktabs}
\usepackage{graphicx}
\usepackage{amssymb}
\usepackage{amsthm}
\usepackage{mathtools}
\usepackage{soul}
\usepackage{subcaption}
\usepackage{algorithm}
\usepackage{algpseudocode}
\usepackage{sourcesanspro} 
\usepackage[OT1]{fontenc}
\usepackage{wrapfig}
\usepackage{colortbl}
\usepackage{titletoc}
\usepackage[most]{tcolorbox}

\newcommand\red[1]{\textcolor{red}{#1}}

\title{\name{}: Model-Intrinsic Reward Alignment for Compositional Diffusion}

\author{%
    Debottam Dutta \thanks{Corresponding author. Email: dd24@illinois.edu} \quad 
    Jianchong Chen \quad 
    Jaehoon Hahm \quad 
    Romit Roy Choudhury
 \\
    University of Illinois at Urbana-Champaign \\
}

\newcommand{\name}{\texttt{TILT}}

\theoremstyle{plain}

\newtheorem{proposition}{Proposition}
\newcommand{\restatedpropref}{}
\newtheorem*{restatedpropinner}{Proposition~\restatedpropref}
\newenvironment{restatedprop}[2]
  {\renewcommand{\restatedpropref}{\ref{#1}}\begin{restatedpropinner}[#2]}
  {\end{restatedpropinner}}
\tcbset{propbox/.style={enhanced, breakable, colback=blue!3!white, colframe=blue!12!white,
  boxrule=0.5pt, arc=2pt, boxsep=0pt, left=5pt, right=5pt, top=5pt, bottom=5pt,
  width=\textwidth, grow to left by=5.5pt, grow to right by=5.5pt, before skip=8pt, after skip=8pt,
  before upper={\csname @arrayparboxrestore\endcsname}}}
\tcolorboxenvironment{proposition}{propbox}
\tcolorboxenvironment{restatedpropinner}{propbox}

\theoremstyle{definition}

\theoremstyle{remark}

\definecolor{co3blue}{rgb}{0.13, 0.23, 0.60}
\definecolor{darkpink}{RGB}{199, 21, 133}
\colorlet{co3first}{co3blue!30!white}
\colorlet{co3second}{co3blue!12!white}
\newcommand{\firstcolor}{\cellcolor{co3first}}
\newcommand{\seccolor}{\cellcolor{co3second}}

\colorlet{co3dark}{co3blue!45!white}
\makeatletter
\newcommand{\algHL}[2]{\hspace*{-\fboxsep}\colorbox{#1}{\makebox[\dimexpr\linewidth-\ALG@thistlm-\fboxsep\relax][l]{#2}}}
\makeatother

\iclrfinalcopy %
\begin{document}

\maketitle
\lhead{Preprint}

\begin{abstract}
    Consider conditional generation $p(x \mid C=\{c_1, c_2, \dots c_k\})$ where $C$ is a prompt composed of multiple concepts $c_i$.
Diffusion models often struggle with compositional prompts, producing samples in which some concepts dominate while others are missing or weakly represented.
Prior work attributes these failures to {\em mode collision}, where single-concept modes of $p(x\mid c_i)$ overlap with modes of the joint $p(x \mid C)$.
To seek out collision-free modes of $p(x \mid C)$, or \emph{pure modes}, corrector-based approaches have attempted to suppress collisions at intermediate diffusion times.
However, local corrections are often heuristic and do not necessarily steer the generation to a \emph{pure mode} in the final data space. Derived from a principled formulation, we present \name{} (\textbf{T}est-time model-\textbf{I}ntrinsic reward a\textbf{L}ignment via \textbf{T}ilting), a training-free framework that poses eventual pure mode sampling as a reward for intermediate-time alignment.
This reward offers valuable advantages: (1) it is intrinsic to the model, hence external reward models need not be trained by modality-specific datasets, (2) it yields a closed-form target under a variational approximation, which makes it realizable through standard diffusion sampling, and (3) it is interpretable, hence amenable to preference-based modifications. \href{https://debottam-dutta7.github.io/tilt_web/}{\textcolor{darkpink}{Project page.}}
\end{abstract}

\section{Introduction}
\label{sec:intro}
Diffusion models have been surprisingly successful across multiple domains, including image ~\citep{sd,imagen,sdxl,t2icompbench}, audio ~\citep{audioldm, musicgen}, video~\citep{sd-video, imagen-video}, molecules, and proteins generation~\citep{alphafold,geodiff,rfdiffusion, diffdock}. These are largely driven by \emph{conditional sampling}, which lets users generate samples that satisfy a given condition, including text-to-image and text-to-audio generation. 
To generate samples that satisfy multiple conditions $c_{1:k}$, a common approach is to compose them into a single condition $C$ and give it to the model as input.
For text-to-image, say, $C$=\texttt{astronaut on moon}, with $c_1$=\texttt{astronaut}, $c_2$=\texttt{on}, and $c_3$=\texttt{moon}.
The model is then expected to generate an image that satisfies all these conditions jointly.

Several studies have observed that diffusion models often suffer from compositional failures when conditioned on multiple concepts; the model confidently drops some concepts or fails to produce a coherent sample that satisfies all of them, \emph{e.g.}, missing objects in text-to-image generation. 
Such failures are broadly described as \emph{concept dominance}: during generation, concepts compete with each other and some dominate the output, so resulting samples miss or only weakly reflect the others.
 
Many approaches exist to mitigate the concept-dominance problem: model surgery approaches manipulate attention maps or model activations~\cite{tome,ae,syngen}; guidance approaches steer sampling trajectories with an external reward models~\cite{dno, ditto}, and correction-based approaches such as \citet{dutta2026steer,tweediemix} correct trajectories with a modified score function. 

Among these, the recent corrector-based approaches have shown promising results. 
~\citet{dutta2026steer} diagnose concept-dominance as \emph{mode collisions} in the model's learned distribution. Under this hypothesis, the model's distribution creates overlaps (or \emph{collisions}) with the modes of individual concept distributions during training. 
During inference, the denoiser's trajectory arrives into one of these overlapping modes and creates concept-dominated samples.
To avoid this, they aim to sample from the \emph{pure modes} of the target distribution by designing a corrector of the form:
\begin{equation}
    p_{t, \text{corrector}}(x) \propto \frac{p_t(x \mid C)} {\prod_{i=1}^{n} p_t(x \mid c_i)}
    \label{eq:intro_corrector}
\end{equation} 

This corrector is used at each intermediate time $t$ during denoising.
The idea is to preserve modes of $p_t(x\mid C)$ so all the concepts contribute jointly, but suppress regions dominated by a single $p_t(x \mid c_i)$ in the denominator. 
This ratio is expected to reduce overlap, resulting in \emph{pure modes}.

Although insightful, these approaches are heuristic. 
They construct the corrector to mitigate overlap in the intermediate distributions (at $t > 0$) while the actual overlap exists in the data-space (at $t=0$).
    \begin{figure}
        \centering
        \vspace{-0.05in}
        \includegraphics[width=\linewidth]{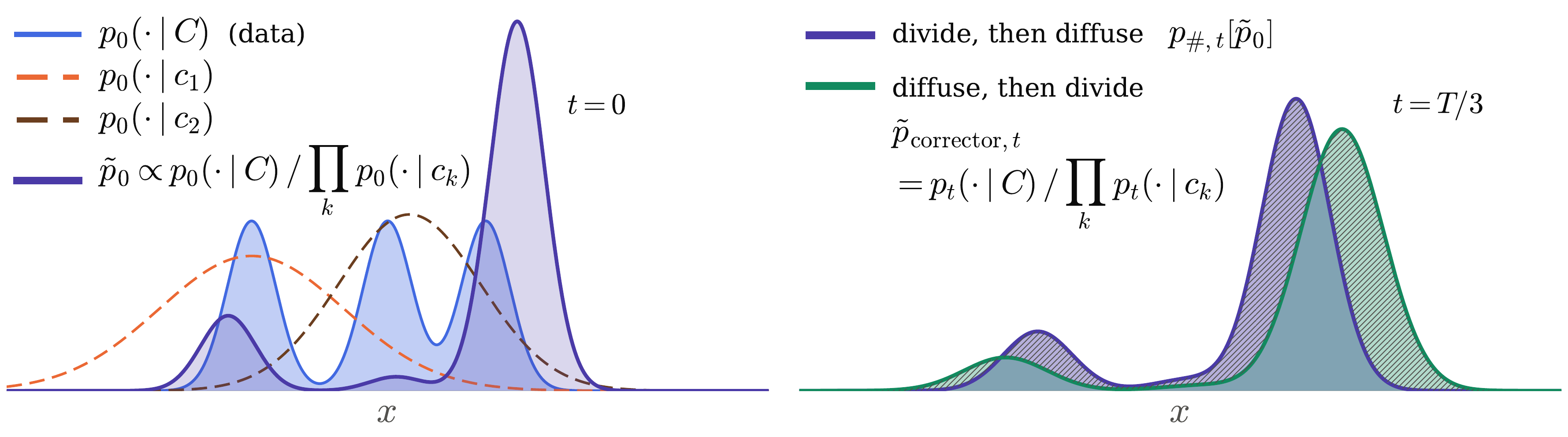}
        \vspace{-0.23in}
        \caption{\small{\textbf{The correct intermediate-time distribution.} \textbf{Left:} in data space ($t=0$),
            the joint $p_0(\cdot \mid C)$ (blue) shares modes with individual concept
            distributions $p_0(\cdot \mid c_1)$ (orange) and $p_0(\cdot \mid c_2)$ (brown). The
            distribution $\tilde p_0$ (purple) describes the \emph{pure modes} -- the modes of the
            joint that no single concept explains. \textbf{Right:} at $t=T/3$, dark-purple is the diffusion pushforward of $\tilde p_0$
            (\emph{divide, then diffuse}), while green is the corrector distribution from Eq. \ref{eq:intro_corrector}, which follows
            \emph{diffuse, then divide}. The two distributions diverge -- a corrector
            defined at $t>0$ therefore steers the sampler towards some other distribution, not the
            noised ``pure mode'' target.}}
\vspace{-0.17in}
        \label{fig:pushforward_compare}
    \end{figure}
As Fig.~\ref{fig:pushforward_compare} shows, noising the true target distribution $\tilde{p}_0$ (purple) to time, say $T/3$, is not the same as noising the individual distributions to $T/3$ and then dividing them to form the corrector $\tilde{p}_t$ (green).
The gap between the purple and green distributions grows significantly with higher number of dimensions, indicating that corrector-based approaches steer their sampler to the wrong target.

We depart from correction-based approaches and instead ask: \emph{"How should intermediate generation be steered so that the denoised samples at $t=0$ are drawn from the pure modes?"} To answer this question, we formulate pure-mode sampling as the terminal reward and derive an intermediate steering direction that maximizes this reward.

Observe that this reward is defined through the model's likelihoods and a diffusion model would not provide it directly\footnote{A naive approach could track the likelihood all the way to a future clean sample, and backpropagate through the entire sampling trajectory, which would be computationally prohibitive.}.
Our key observation is that the reward does not need these likelihoods individually. 
The reward is a ratio of likelihoods, so under the variational approximation of the diffusion likelihood, the intractable terms cancel and the reward reduces to a contrast of denoising errors.
We show how the new reward is interpretable, and can be mapped to a \emph{Diffusion classifier} score that samples from \emph{pure modes} by trying to increase the classification score for the joint condition $C$, while reducing the scores for the individual concept conditions $c_i$.
Finally, since the reward is purely distribution-based, it's also modality-agnostic and does not rely on training an external reward model. 
We show results in both image and audio generation benchmarks demonstrating that such a reward is effective for compositional generation and viable in both domains. 
Our contributions are: 

\circled{1} We derive a tractable reward for intermediate steps of the denoising process that provably optimizes the end-goal of sampling \emph{pure modes}. This reward is intrinsic to the model, hence generalizes across modalities (obviating the need for data-trained reward models); it is easily interpretable.

\circled{2} We show why the reward works:  maximizing the reward is equivalent to maximizing the total correlation between a sample and the input conditions.  Intuitively, the reward favors samples that satisfy the synergy or coherence among the concepts.

\circled{3} Results show how the reward consistently avoids mode overlap or concept dominance by trying to select samples that increase the classification score for condition $C$ while reducing the scores for the individual concepts $c_{1:k}$.

\section{Background}
\label{sec:background}
\vspace{-0.1in}

$\blacksquare$ \textbf{Diffusion Sampling and Classifier-Free Guidance.}
Diffusion models~\citep{ddpm,song2021scorebased} learn to reverse a noising process that maps data $x_0 \sim p_0$ to noise $x_T \sim p_T$. Conditional models approximate the score $\nabla_{x_t}\log p_t(x_t\mid c)$, commonly parameterized by a noise prediction $\epsilon_\theta(x_t,t\mid c)$. Given $x_t$, the Tweedie estimate of the clean sample is
\begin{equation}
    \hat{x}_0 =
    \frac{x_t-\sqrt{1-\alphabar_t}\,
    \epsilon_\theta(x_t,t\mid c)}
    {\sqrt{\alphabar_t}}.
    \label{eq:tweedie}
\end{equation}
A deterministic DDIM step then updates
$
    x_{t-1}
    =
    \sqrt{\alphabar_{t-1}}\,\hat{x}_0
    +
    \sqrt{1-\alphabar_{t-1}}\,
    \epsilon_\theta(x_t,t\mid c)
    \label{eq:ddim}
$
~\citep{ddim}. For text-conditioned generation, classifier-free guidance (CFG)~\citep{cfg}
combines conditional and unconditional noise predictions as
\begin{equation}
    \epsilon_t^{\lambda,c}
    =
    \lambda\,\epsilon_\theta(x_t,t\mid c)
    +(1-\lambda)\,\epsilon_\theta(x_t,t\mid\varnothing),
    \label{eq:cfg}
\end{equation}
and substitutes $\epsilon_t^{\lambda,c}$ into the sampling update.
CFG++~\citep{chung2024cfg++} instead uses the guided prediction to estimate
$\hat{x}_0$ while retaining the unconditional prediction for the noise
direction:
\begin{equation}
    x_{t-1}^{\mathrm{CFG++}}
    =
    \sqrt{\alphabar_{t-1}}\,
    \hat{x}_0[\epsilon_t^{\lambda,c}]
    +
    \sqrt{1-\alphabar_{t-1}}\,
    \epsilon_\theta(x_t,t\mid\varnothing).
    \label{eq:cfgpp}
\end{equation}

We build on this clean-space view of diffusion sampling, as our method
modifies the Tweedie estimate $\hat{x}_0$ using a model-intrinsic reward
before propagating the next sampling step.

$\blacksquare$ \textbf{Diffusion Steering with Rewards.}
Diffusion sampling can also be steered by an external reward $r(x_0)$, e.g., a human-preference or aesthetic score \citep{kim2025das}. Such rewards are typically defined only on clean samples $x_0$. Given a pre-trained model with distribution $p_{\mathrm{pre}}$, the goal is a target distribution that maximizes expected reward while staying close to $p_{\mathrm{pre}}$:
\begin{equation}
    p_{\mathrm{tar}} = \arg\max_{p} ~ [ \mathbb{E}_{x_0 \sim p}[r(x_0)]] \quad \text{s.t.} \quad D_{\mathrm{KL}}(p \,\|\, p_{\mathrm{pre}}) < \epsilon.
    \label{eq:reward-hard}
\end{equation}
Introducing a Lagrange multiplier $\alpha > 0$ for the constraint, \eqref{eq:reward-hard} can equivalently be written in its unconstrained, penalized form:
\begin{equation}
p_{\mathrm{tar}} = \arg\max_{p}
\left[
\mathbb{E}_{x_0 \sim p}[r(x_0)]
- \alpha D_{\mathrm{KL}}(p \,\|\, p_{\mathrm{pre}})
\right].
\label{eq:reward-kl}
\end{equation}
Taking the functional derivative of this objective with respect to $p$ and enforcing the normalization constraint $\int p(x_0)dx_0=1$ yields the optimality condition
$\log p_{\mathrm{tar}}(x_0)
= \log p_{\mathrm{pre}}(x_0) + r(x_0)/\alpha + \mathrm{const}$.
Thus, the regularized objective admits the closed-form solution as follows, where $Z$ is the normalizing constant ensuring that $p_{\mathrm{tar}}$ integrates to one.
\begin{equation}
p_{\mathrm{tar}}(x_0)
= \frac{1}{Z}
p_{\mathrm{pre}}(x_0)
\exp{(\frac{r(x_0)}{\alpha}}),
\qquad
Z
= \int
p_{\mathrm{pre}}(x_0)
\exp\left(\frac{r(x_0)}{\alpha}\right)dx_0
\label{eq:reward-target}
\end{equation}

\section{Diffusion Steering with Intrinsic Reward}
\vspace{-0.1in}
We first motivate with a simple experiment that compositional failures arise from overlap modes. Then we formulate pure-mode sampling as reward alignment and show theoretical justification as to why the model-intrinsic reward is a readily available and useful quantity for alignment. We end with the derived reward guidance algorithm.

\subsection{Pure mode as an Intrinsic reward}
\label{sec:m2-formulation}
\begin{wrapfigure}{r}{0.35\linewidth}
    \vspace{-0.2in}
    \centering
    \includegraphics[width=\linewidth]
    {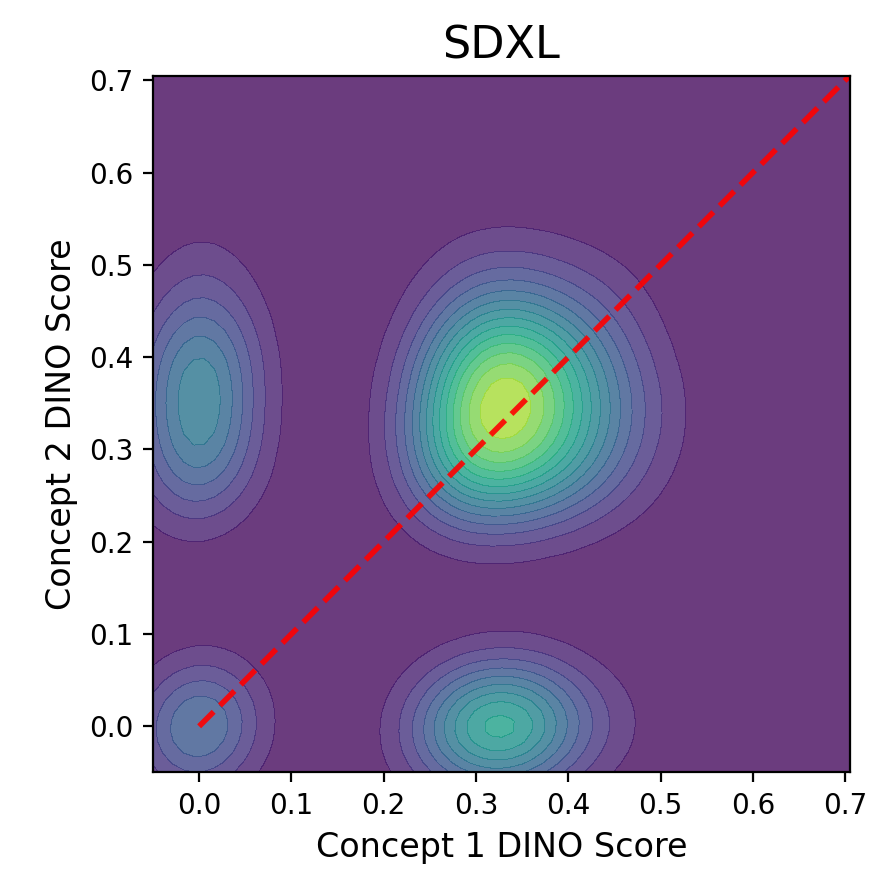}
    \caption{Density plot of DINOv2 scores between each sub-prompt $c_1, c_2$ and the generated image using SDXL. This visualizes the failure in compositional generation.}
    \label{fig:concept_dominance}
    \vspace{-0.1in}
\end{wrapfigure}

$\blacksquare$ \textbf{Observation on compositional generation.} 
We start with an experiment by giving prompts of the form ``\texttt{a [Animal] and a [Object]}'' to SDXL~\citep{sdxl} and generated samples over multiple seeds. Figure~\ref{fig:concept_dominance} measures DINOv2 concept-quality scores for generated samples from these compositional prompts. 
A significant portion of the empirical density concentrates near the axes $x{=}0$ and $y{=}0$, indicating that conditional samples disproportionately retain only one of the two concepts. Crucially, the modes near the axes are precisely the regions where the joint conditional density $p^\theta(x_0 \mid C)$ overlaps with one of the marginals $p^\theta(x_0 \mid c_i)$. This substantiates the hypothesis that compositional failures are closely linked to mode overlap.

We want a sampling target whose mass concentrates on \emph{pure} modes and avoids overlap modes. A natural construction reweights the joint by the inverse product of marginals:
\begin{equation}
\tilde p(x_0 \mid C) \;\propto\; \frac{p^\theta(x_0 \mid C)}{\prod_{i=1}^{K} p^\theta(x_0 \mid c_i)}.
\label{eq:m2-tildep}
\end{equation}
The reweighting suppresses regions where any single $p^\theta(x \mid c_i)$ is large and preserves regions with concept synergy where mass under the joint is supported by all concepts in a balanced way. 
Notably, the above distribution lies in data space or over $x_0$. As discussed in \ref{sec:intro}, previous works \citet{dutta2026steer, tweediemix} target sampling from the above distribution by defining a heuristic corrector of the same form at all $t>0$.

$\blacksquare$ \textbf{Formulation.} We break away from the heuristic corrector approach and we pose the "pure mode" sampling as test-time \emph{reward alignment} problem at $t=0$. 
Following \eqref{eq:reward-hard}, we set up our optimization as: 
\begin{equation}
\begin{aligned}
p^* \;=\; \argmax_{p}\; & \E_{x_0 \sim p}\!\Big[\log \tfrac{p^\theta(x_0 \mid C)}{\prod_{i=1}^{K} p^\theta(x_0 \mid c_i)^{w_i}}\Big] 
\quad \text{s.t.} & \mathrm{D}_\mathrm{KL} \big(p \,\Vert\, p^\theta(\cdot \mid C)\big) < \epsilon.
\end{aligned}
\label{eq:m2-rl}
\end{equation}
where $w_i \in \R^+$ and $\sum_i w_i = 1$. Note that, the reward $R_w(x_0):= \log \frac{p^\theta(x_0 \mid C)}{\prod_{i=1}^{K} p^\theta(x_0 \mid c_i)^{w_i}}$ is defined intrinsically from the pretrained model itself. 
The Proposition below theoretically justifies why the optimization in \eqref{eq:m2-rl} leads to a distribution that emphasizes pure modes where all concepts synergize.
\begin{proposition}[Pure-mode reward and concept synergy, informal]
\label{prop:m2-tc}
\noindent (i) For a fixed prompt $C$, applying Bayes' rule to the reward $R_w$
in~\eqref{eq:m2-rl} gives, up to a constant independent of $x_0$,
\begin{equation}
R_w(x_0) \;=\;
\underbrace{\log \frac{p^\theta(C \mid x_0)}{\prod_{i} p^\theta(c_i \mid x_0)}}_{\text{point-wise conditional Total Correlation (pcTC) } \iota_C(x_0)}
\;+\;
\underbrace{\sum_{i} (1 - w_i) \log p^\theta(c_i \mid x_0)}_{\text{per-concept evidence } \pi_w(x_0)}
\;+\; \mathrm{const}.
\label{eq:m2-tc}
\end{equation}
Hence solving~\eqref{eq:m2-rl} amounts to maximizing the expected pcTC plus per-concept evidence
while staying close to $p^\theta(\cdot \mid C)$, and the solution $p^*$ raises this quantity
relative to the pretrained model.

\noindent (ii) At the
population level (across prompts and samples),
\begin{equation}
\E_{C}\,\E_{x_0}\big[R_w(x_0)] \;=\;
\underbrace{\mathrm{TC}(c_1; \dots; c_K \mid X_0)}_{\text{conditional Total Correlation}}
\;-\;
\underbrace{\sum_{i} (1 - w_i)\, H(c_i \mid X_0)}_{\text{per-concept uncertainty}} \ + \ \mathrm{const},
\label{eq:m2-tc-pop}
\end{equation}
where $\mathrm{TC}(c_1; \dots; c_K \mid X_0) = \sum_i H(c_i \mid X_0) - H(c_1, \dots, c_K \mid X_0)$ denotes
the conditional Total Correlation~\citep{watanabe_tc}, i.e., how strongly the concepts explaining a
generated sample depend on each other, and $H(c_i \mid X_0)$ denotes the uncertainty about concept $c_i$
that remains after seeing the sample (Appendix~\ref{app:tc-derivation}).
\end{proposition}
$\blacksquare$ \noindent\textbf{Remarks.}
\circled{1} \emph{Synergy.} $\iota_C$ is the pointwise conditional total correlation (pcTC), the
per-sample analogue of conditional Total Correlation~\citep{watanabe_tc}. For $K = 2$ it equals
pointwise mutual information between $c_1$ and $c_2$ given $x_0$. It is high when the model's
posterior at $x_0$ explains the concepts \emph{jointly} rather than independently, and it vanishes
when the posterior factorizes over concepts.
\circled{2} \emph{Presence.} $\pi_w$ rewards the evidence $x_0$ provides for each concept on its own,
so synergy cannot be gained by depicting every concept only weakly.
\circled{3} \emph{Pure modes.} Together, $\iota_C + \pi_w = \sum_i w_i \log \frac{p^\theta(C \mid x_0)}{p^\theta(c_i \mid x_0)}$,
which is negative when a single concept $c_i$ satisfies $x_0$ better than the
full prompt $C$. These are the overlap modes of Fig.~\ref{fig:concept_dominance}, which $p^*$
down-weights. 
\circled{4} \emph{Population view.} At the population level, the pure-mode objective favors high Total
Correlation among the concepts while penalizing concepts that cannot be identified from the sample.

\subsection{Sampling using Pure Mode Guidance}
\label{sec:pure-mode-guidance}

As discussed in \secref{sec:background}, the problem in \eqref{eq:m2-rl} admits the closed form solution:  
\begin{equation}
\begin{aligned}
p^*(x_0) \;&=\; \tfrac{1}{Z}\, p^\theta(x_0 \mid C)\, \exp\!\big(\beta\, R_w(x_0)\big)
\end{aligned}
\label{eq:m2-pstar}
\end{equation}
i.e., $\log p^*(x_0) \;=\; \log p^\theta(x_0 \mid C) + \beta\, R_w(x_0) - \log Z$, where $R_w(x_0) = \log \tfrac{p^\theta(x_0 \mid C)}{\prod_i p^\theta(x_0 \mid c_i)}$.

$\blacksquare$ \textbf{Reward alignment as posterior sampling.}
We introduce a binary optimality variable $O\in\{0,1\}$ with
\begin{equation}
p(O=1\mid x_0,C)
\propto
\exp\!\big(\beta R_w(x_0)\big),
\end{equation}

such that samples with higher intrinsic reward are exponentially more likely to be
optimal. Applying Bayes' rule with prior $p^\theta(\cdot\mid C)$ gives
\begin{equation}
p(x_0\mid C,O=1)
\propto
p^\theta(x_0\mid C)
\exp\!\big(\beta R_w(x_0)\big)
\propto p^*(x_0).
\end{equation}

Thus, sampling from the reward-tilted distribution $p^*$ is equivalent
to posterior sampling conditioned on optimality. We omit ``$O=1$''
hereafter for brevity.
Reverse diffusion requires the posterior score at every diffusion time:
\begin{equation}
\nabla_{x_t}\log p(x_t\mid C,O)
=
\underbrace{\nabla_{x_t}\log p(x_t\mid C)}
_{\text{prior score}}
+
\underbrace{\nabla_{x_t}\log p(O\mid x_t,C)}
_{\text{reward guidance}}.
\label{eq:m2-master}
\end{equation}
The latter term is generally intractable since
\begin{equation}
p(O\mid x_t,C)
=
\E_{p(x_0\mid x_t,C)}
\left[p(O\mid x_0,C)\right].
\end{equation}

Following posterior-guided diffusion \citep{dps}, we approximate this expectation
using the Tweedie's formula for posterior mean $\hat{x}_0(x_t)$:
\begin{equation}
p(O\mid x_t,C)
\approx
p(O\mid\hat{x}_0(x_t),C).
\end{equation}

Substituting the optimality likelihood gives
\begin{equation}
\nabla_{x_t}\log p(O\mid x_t,C)
\approx
\beta\nabla_{x_t}\hat R_w(x_t),
\qquad
\hat R_w(x_t)
\coloneq
R_w(\hat{x}_0(x_t)).
\label{eq:m2-approx-guidance}
\end{equation}

$\blacksquare$  \textbf{Approximating the model-intrinsic reward.}
Direct evaluation of $R_w(\hat{x}_0)$ requires the terminal likelihoods
$p^\theta(\hat{x}_0\mid C)$ and
$p^\theta(\hat{x}_0\mid c_i)$.
Our key observation is that the reward can be written as a sum of
log-likelihood ratios,
\begin{equation}
R_w(\hat{x}_0)
=
\sum_{i=1}^{K}
w_i
\log
\frac{
p^\theta(\hat{x}_0\mid C)
}{
p^\theta(\hat{x}_0\mid c_i)
}.
\label{eq:reward-ratio}
\end{equation}

Using the variational form of the diffusion log-likelihood,
$-\log p^\theta(x_0\mid c)
=
\int_0^T
\gamma(t)
\E_{\epsilon}
\left[
\left\|
\epsilon-
\epsilon_\theta(x_t,t\mid c)
\right\|^2
\right]
\diff t
+
\mathrm{const},
$
we can approximate the likelihood ratio using the differences of Diffusion
Classifier Energies \citep{diff-classifier}:

\begin{equation}
\hat R_w(\hat{x}_0)
=
-\frac{1}{K}
\sum_{i=1}^{K}
w_i
\left[
E_C(\hat{x}_0)
-
E_{c_i}(\hat{x}_0)
\right],
\label{eq:energy_reward}
\end{equation}
where
$
E_c(x_0)
=
\E_{\tau,\epsilon}
\left[
\gamma(\tau)
\left\|
\epsilon-
\epsilon_\theta(x_\tau,\tau\mid c)
\right\|^2
\right].
\label{eq:diffusion-energy}
$
Here,
$x_\tau=
\sqrt{\bar\alpha_\tau}x_0+
\sqrt{1-\bar\alpha_\tau}\epsilon$.
We use the same sampled $(\tau,\epsilon)$ across the joint and
constituent conditions when estimating their energy differences. We want to minimize the energy $E_C$ while maximizing energies $E_{c_i}$. 
The full derivation and interpretation are provided in
Appendix~\ref{app:reward-derivation}.

$\blacksquare$ \textbf{Reward-corrected Tweedie sampling.}
\Eqref{eq:m2-approx-guidance} suggests differentiating the
reward through the Tweedie estimate, $ \nabla_{x_t}\hat R_w(x_t)
=
J_{\hat{x}_0}(x_t)^\top
\nabla_{\hat{x}_0}
\hat R_w(\hat{x}_0), $
where $J_{\hat{x}_0}(x_t)$ denotes the Jacobian of the Tweedie
prediction with respect to $x_t$.
Rather than explicitly propagating the reward gradient through this
Jacobian, we use a simple Jacobian-free clean-space corrector.
At each denoising step, we compute $\nabla_{\hat{x}_0} \hat R_w(\hat{x}_0)$ and directly update the predicted clean sample as
\begin{equation}
\tilde{x}_0
\gets
\hat{x}_0
+
\beta \ \nabla_{\hat{x}_0} \hat R_w(\hat{x}_0)
\label{eq:m2-opt1}
\end{equation}

where $\beta$ is the reward strength.
The corrected prediction $\tilde{x}_0$ is then used in place of
$\hat{x}_0$ in the subsequent DDIM update, following \citet{mpgd}. Note that this does not require external reward model or fine-tuning of the model.

\begin{algorithm*}[t]
\caption{\name{}: Diffusion sampling using model-intrinsic reward}
\label{alg:m2-hybrid}
\begin{algorithmic}[1]

\Require pretrained diffusion model $\epsilon_\theta$,
multi-prompt $C=\{c_i\}_{i=1}^K$, number of diffusion steps $N$,
stopping index $M$, schedule $\{t_n\}$,
reward strength $\beta$.

\State $x_T \sim \mathcal{N}(0,I)$

\For{$n=N,N{-}1,\ldots,M+1$}

    \State \algHL{co3second}{$\hat{x}_0 \gets
    \dfrac{
    x_{t_n}
    -\sqrt{1-\bar{\alpha}_{t_n}}\,
    \epsilon_\theta(x_{t_n},t_n\mid C)}
    {\sqrt{\bar{\alpha}_{t_n}}}$
    \Comment{Tweedie estimate}}

    \State \algHL{co3first}{$\hat R_t \gets
    \dfrac{1}{K}
    \sum_{i=1}^{K}
    w_i
    \left[
    E_C(\hat{x}_0)-E_{c_i}(\hat{x}_0)
    \right]$
    \Comment{Energy-based reward, Eq.~\ref{eq:energy_reward}}}
    
    \State \algHL{co3dark}{$\tilde{x}_0
    \gets
    \hat{x}_0+\beta \nabla_{\hat{x}_0}\hat R_t$
    \Comment{Reward correction, Eq.~\ref{eq:m2-opt1}}}

    \State \setlength{\lineskip}{0pt}%
    \algHL{co3dark}{$x_{t_{n-1}}
    \gets
    \mathrm{DDIM}\!\left(
    \tilde{x}_0;
    \epsilon_\theta(x_{t_n},t_n\mid C)
    \right)$}
    \vspace{0.01in}

\EndFor

\State \Return $x_0$

\end{algorithmic}
\end{algorithm*}
\section{Experiments}
\vspace{-0.1in}
$\blacksquare$ \textbf{Implementation Details.}
For experiments in image space, we use SDXL~\citep{sdxl} as a base model and without any additional training or fine-tuning. Our method modifies only the inference procedure. We use the DDIM scheduler with CFG (scale $5.0$) with 50 denoising steps. 
In text2audio space, we use AudioLDM~\citep{audioldm} as a base model with DDIM with CFG (scale $3.5$) as the base sampler with $50$ steps. 
The proposed correction is applied on early denoising steps.
Unless otherwise specified, we correct the first $5$ denoising steps, use $2$ latent correction steps per corrected timestep, and use $5$ correction iterations at the initial timestep. More implementation details in Appendix~\ref{app:implementation}. 

\begin{wraptable}{r}{0.34\linewidth}
\vspace{-12pt}
\small
\centering
\caption{Quantitative results on GenEval prompts.}
\label{tab:quant_geneval}
\begin{tabular}{lcc}
\toprule
Method & T2V $\uparrow$ & HPSv3 $\uparrow$ \\
\midrule
CFG & 0.3934 & 5.5029 \\
CFG++ & \seccolor 0.5073& \seccolor 5.9806 \\
R2F & 0.4799 &  5.8176 \\
SuperDiff & 0.1589 & 0.082 \\
CO3 & 0.5014 & 4.3026 \\
DOS & 	0.5025 & 5.5744 \\
TILT & \firstcolor 0.5192 & \firstcolor 6.1219 \\
\bottomrule
\end{tabular}
\vspace{-8pt}
\end{wraptable}

$\blacksquare$ \textbf{Evaluation benchmarks and metrics.}
For text-to-image generation, we use \textbf{GenEval}~\citep{ghosh2023geneval} for simple prompts, an object-centric benchmark covering co-occurrence, count, color, and spatial relations, and \textbf{T2I-CompBench}~\citep{huang2023t2icompbench} for open-world prompts, evaluating its \textit{Color}, \textit{Shape}, \textit{Texture}, and \textit{Complex} subsets.

For images, we report \textbf{T2V}~\citep{t2v} for text–image alignment and \textbf{HPSv3}~\citep{ma2025hpsv3} for learned human preference. For audio, we report \textbf{CLAP}~\citep{elizalde2023clap} for text–audio alignment; \textbf{FAD}~\citep{kilgour2019frechet} and \textbf{FD}~\citep{audioldm}, which measure the distance between generated and reference audio features extracted with VGGish and PANNs, respectively; and \textbf{KL} divergence between their predicted audio-event distributions. Higher is better for T2V, HPSv3, and CLAP; lower is better for FAD, FD, and KL. CLIP~\citep{clip} is used only as an external reward in baseline comparisons. To test whether our reward generalizes beyond images, we also evaluate text-to-audio generation on \textbf{AudioCaps}~\citep{kim2019audiocaps}, a dataset of audio clips paired with human-written captions.

$\blacksquare$ \textbf{Comparison baselines.}
For text-to-image generation, we compare against \textbf{CFG}~\citep{cfg},
\textbf{CFG++}~\citep{chung2024cfg++}, and prior compositional generation
methods, including \textbf{SuperDiff}~\citep{superdiff},
\textbf{R2F}~\citep{r2f}, \textbf{CO3}~\citep{dutta2026steer}, and
\textbf{DOS}~\citep{dos}, using the same SDXL~\citep{podell2023sdxl}
backbone. For text-to-audio generation, we use \textbf{AudioLDM}~\citep{audioldm}
as the base generator and compare against the reward-steering method
\textbf{SCORE}~\citep{score}. Unless otherwise specified, all comparisons use
the same underlying pretrained model and evaluation setting within each modality (see Sec.~\ref{sec:related} for more details).

\subsection{Comparison on Text-to-Image Generation}
$\blacksquare$ \textbf{Simple Prompts.} 
Table~\ref{tab:quant_geneval} shows the results for \name{} on GenEval prompts. All the methods use SDXL as a base model and do not use or depend on any external signal for generation. All the baselines here depend only on the base model and don't utilize any external reward signal like \name{}. In both metrics \name{} outperforms the baselines. Especially, our method gains significantly over the baselines in HPSv3 metric which aligns with human preferences. Notably, it also outperforms CO3 which is a corrector-based approach that uses a corrector distribution that is structurally similar to \eqref{eq:reward-kl}. The result validates our claim that "pure mode" as a data-space reward is more effective than intermediate-heuristic correctors. 
\begin{table}[t]
\centering
\small
\setlength{\tabcolsep}{4pt}
\caption{Quantitative results on T2I-CompBench prompts.}
\label{tab:quant_t2icomp_full}
\begin{tabular}{lcccccccc}
\toprule
& \multicolumn{2}{c}{Color} & \multicolumn{2}{c}{Shape} & \multicolumn{2}{c}{Texture} & \multicolumn{2}{c}{Complex} \\
\cmidrule(lr){2-3} \cmidrule(lr){4-5} \cmidrule(lr){6-7} \cmidrule(lr){8-9}
Method & T2V $\uparrow$ & HPSv3 $\uparrow$ & T2V $\uparrow$ & HPSv3 $\uparrow$ & T2V $\uparrow$ & HPSv3 $\uparrow$ & T2V $\uparrow$ & HPSv3 $\uparrow$ \\
\midrule
CFG
& 0.5183 
& 5.1216 
& 0.3565 
& 6.1377 
& 0.5529 
& 6.5021 
& 0.6238 
& 7.5702 \\

CFG++
& 0.5668 
& \seccolor 5.7358 
& 0.3839 
& \seccolor 6.9679 
& 0.5839 
& \seccolor 7.1200 
& 0.6307 
& \seccolor 8.1785 \\

R2F 
& 0.5328 
& 5.5146 
& 0.3452 
& 6.5545 
& 0.5573 
& 6.7431 
& 0.6234 
& 8.0638 \\

CO3 
& \firstcolor 0.5938 
& 4.4736 
& \seccolor 0.3895 
& 6.1982 
& 0.4252 
& 6.2801 
& 0.6313 
& 6.6193 \\

DOS 
& 0.5028 
& 4.8979 
& 0.3865 
& 6.0388 
& \firstcolor 0.6199 
& 6.4046 
& \firstcolor 0.6603 
& 6.4046 \\

SuperDiff 
& 0.2420 
& -0.2098 
& 0.1660 
& 2.6294 
& 0.2528 
& 0.4163 
& 0.3865 
& 2.3976 \\

\name{} (Ours)
& \seccolor 0.5688 
& \firstcolor 5.8002 
& \firstcolor 0.4019 
& \firstcolor 7.0523 
& \seccolor 0.5932 
& \firstcolor 7.2136 
& \seccolor 0.6350 
& \firstcolor 8.2156 \\

\bottomrule
\end{tabular}
\end{table}

$\blacksquare$ \textbf{Complex Prompts.} Table~\ref{tab:quant_t2icomp_full} shows \name{}'s performance on more complex scenarios using the T2ICompbench prompts. Our method consistently outperforms or matches the performance of the baselines across most of the evaluated metrics. Similar to simpler prompts in Table \ref{tab:quant_geneval}, \name{} also demonstrates superior performance in capturing the nuances of complex compositions and validates that our proposed intrinsic reward is equally effective in the more complex and challenging scenarios. Our method is the best performing in HPSv3 metric that mimics human preferences. 

\begin{figure}[t]
    \centering
    \newlength{\qualImgW}\newlength{\qualLabW}\newlength{\qualGap}%
    \setlength{\qualGap}{1pt}%
    \setlength{\qualLabW}{28pt}%
    \setlength{\qualImgW}{\dimexpr(\linewidth-\qualLabW-7\qualGap)/7\relax}%
    \newcommand{\qualFont}{\fontencoding{T1}\sffamily}%
    \newcommand{\qualVC}[1]{$\vcenter{\hbox{#1}}$}%
    \newcommand{\qualHead}[1]{\hspace{\qualGap}\makebox[\qualImgW]{\qualFont\fontseries{sb}\fontsize{7}{8}\selectfont #1}}%
    \newcommand{\qualPrompt}[1]{\makebox[\qualLabW]{\qualVC{\rotatebox{90}{\parbox{\qualImgW}{\centering\qualFont\itshape\fontsize{6.5}{7}\selectfont #1}}}}}%
    \newcommand{\qualImg}[1]{\hspace{\qualGap}\qualVC{\includegraphics[width=\qualImgW]{figs/qualitative_grid/#1.jpg}}}%
    \newcommand{\qualRow}[2]{\qualPrompt{#2}\qualImg{#1_sdxl}\qualImg{#1_sdxlcfgpp}\qualImg{#1_co3}\qualImg{#1_superdiff}\qualImg{#1_dos}\qualImg{#1_r2f}\qualImg{#1_ours}\par}%
    {\offinterlineskip
    \makebox[\qualLabW]{}\qualHead{SDXL}\qualHead{SDXL + CFG++}\qualHead{CO3}\qualHead{SuperDiff}\qualHead{DOS}\qualHead{R2F}\qualHead{\fontseries{b}\selectfont TILT (Ours)}\par
    \vspace{2pt}
    \qualRow{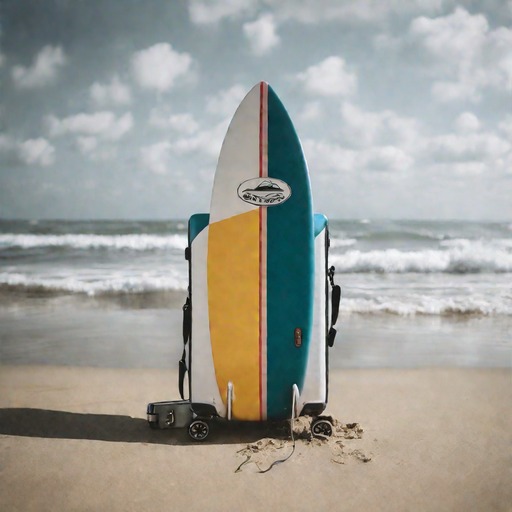}{a photo of a\\surfboard\\and a suitcase}\vspace{\qualGap}
    \qualRow{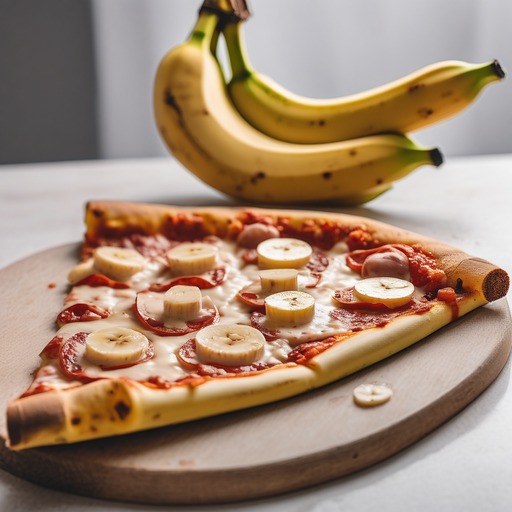}{a photo of a pizza\\right of a banana}\vspace{3pt}
    \qualRow{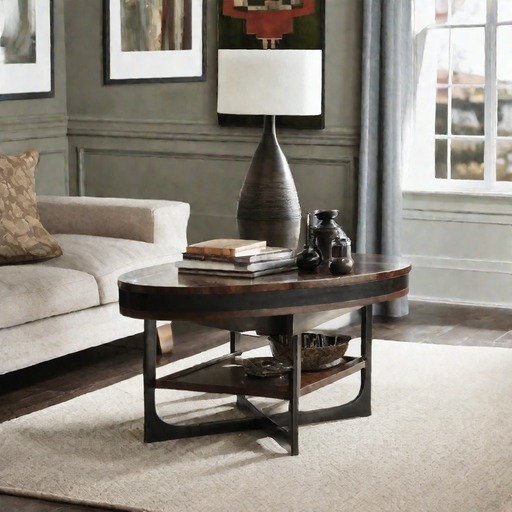}{an oval coffee table\\and a square\\end table}\vspace{\qualGap}
    \qualRow{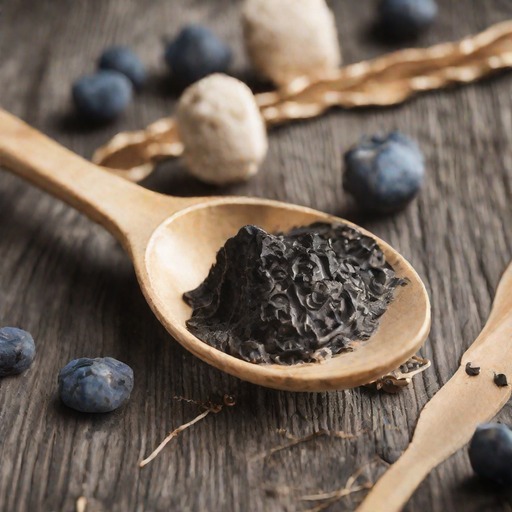}{a metallic\\jewelry and a\\wooden spoon}\vspace{\qualGap}
    \qualRow{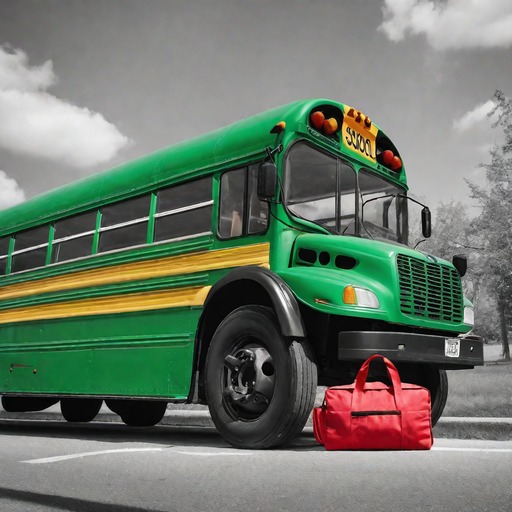}{a green school bus\\and a red bag}\vspace{\qualGap}
    \qualRow{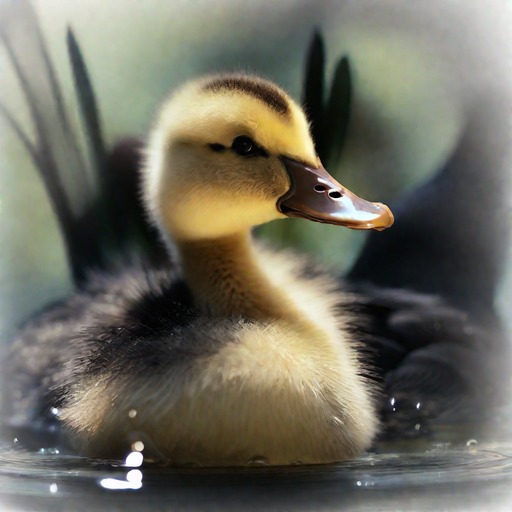}{The soft yellow\\duckling swam\\next to the\\sleek black swan.}
    }
    \caption{\textbf{Qualitative comparison on GenEval and T2I-CompBench.} The top two rows use prompts from GenEval and the bottom four rows use prompts from T2I-CompBench (shape, texture, color, and complex subsets); each row's prompt is shown on the left. All methods use the SDXL backbone and share the same random seed within each row.}
    \label{fig:qualitative}
    \vspace{-10pt}
\end{figure}

\begin{wraptable}{r}{0.5\linewidth}
\vspace{-12pt}
\small
\centering
\caption{Comparison with extrinsic-reward steering on GenEval.}
\label{tab:quant_vision_extrinsic}
\vspace{-0.05in}
\begin{tabular}{lccc}
\toprule
Method  & BLIP-VQA $\uparrow$ & T2V $\uparrow$ & HPSv3 $\uparrow$ \\
\midrule
SDXL &  0.339 & 0.483 & 5.50 \\
\midrule
FKS-CLIP  &  \firstcolor 0.450 & \firstcolor 0.563 & 5.89 \\
FKS-\name{}  &  0.446 & 0.530 & \firstcolor 5.91 \\
\bottomrule
\end{tabular}
\vspace{-8pt}
\end{wraptable}
$\blacksquare$ \textbf{Qualitative Comparison.} Figure~\ref{fig:qualitative} compares \name{} qualitatively with other baselines. The first two rows use GenEval prompts. SDXL, CFG++ and CO3 struggled with simpler prompts of the form \texttt{"a [concept1] and a [concept2]"} (1st row). With a slightly complicated modifier ``right of'' (2nd row), most methods were unable to generate the desired image. Bottom four rows show results on more complicated scenarios from T2ICompBench. $3$rd and $4$th rows contain \textit{shape} and \textit{texture} information while $5$th row has \textit{color} modifiers. Most methods either drop concepts (row 4) or mix the attributes between concepts (row 5). Last row contains \textit{complex} prompts that combine multiple concepts with relations between them. \name{} retains better fidelity even in these challenging cases.
\vspace{-0.1in}
\subsection{Comparison with methods using extrinsic rewards} 
$\blacksquare$ \textbf{Text-to-Image Generation.} We compare against FKS~\citep{fksteering}, a sequential Monte Carlo method that steers diffusion with an external CLIP reward. It runs $4$ particles, scores them with CLIP at intermediate steps, and resamples to keep the most promising ones. To isolate the effect of the reward, we keep the FKS sampler fixed and replace CLIP with our intrinsic reward. Table~\ref{tab:quant_vision_extrinsic} shows that FKS-\name{} performs on par with FKS-CLIP and surpasses it on HPSv3. This comparison favors CLIP: it is an external text-image alignment model trained on a large-scale corpus of paired text and images, so it brings supervision from outside the generator. \name{}, by contrast, is computed from the base diffusion model and uses no external data, annotations, or auxiliary networks. That a reward requiring no outside supervision is competitive with a large externally trained reward model shows that the base model already contains much of the signal needed to correct compositional failures, and that \name{} makes this signal usable without collecting data or training an external model. Figure~\ref{fig:fks_main} illustrates the qualitative comparison. More visualizations in Appendix~\ref{app:vis_qualitiative}.\\
\begin{figure}[t]
    \centering
    \begin{minipage}[t]{0.42\linewidth}
    \vspace{0pt}\centering
    \ifdefined\fksImgW\else\newlength{\fksImgW}\newlength{\fksLabW}\newlength{\fksGap}\newlength{\fksTotW}\fi%
    \setlength{\fksGap}{1pt}%
    \setlength{\fksLabW}{28pt}%
    \setlength{\fksTotW}{\linewidth}%
    \setlength{\fksImgW}{\dimexpr(\fksTotW-\fksLabW-3\fksGap-0.5pt)/3\relax}%
    \newcommand{\fksFont}{\fontencoding{T1}\sffamily}%
    \newcommand{\fksVC}[1]{$\vcenter{\hbox{#1}}$}%
    \newcommand{\fksHead}[1]{\hspace{\fksGap}\makebox[\fksImgW]{\fksFont\fontseries{sb}\fontsize{7}{8}\selectfont #1}}%
    \newcommand{\fksPrompt}[1]{\makebox[\fksLabW]{\fksVC{\rotatebox{90}{\parbox{\fksImgW}{\centering\fksFont\itshape\fontsize{6.5}{7}\selectfont #1}}}}}%
    \newcommand{\fksImg}[1]{\hspace{\fksGap}\fksVC{\includegraphics[width=\fksImgW]{#1.jpg}}}%
    \newcommand{\fksRow}[3]{\fksPrompt{#3}\fksImg{#1/cfg/#2}\fksImg{#1/fks_clip_bestof4/#2}\fksImg{#1/fks_co3r_bestof4/#2}\par}%
    {\offinterlineskip
    \makebox[\fksLabW]{}\fksHead{SDXL}\fksHead{FKS-CLIP}\fksHead{\fontseries{b}\selectfont FKS-\name{}}\par
    \vspace{2pt}
    \fksRow{figs/qualitative/fks}{08_s1_a_photo_of_a_green_teddy_bear_and_a_brown_kite}{a photo of a\\green teddy bear\\and a brown kite}\vspace{\fksGap}
    \fksRow{figs/qualitative/fks}{01_s1_a_photo_of_a_black_car_and_a_green_parking_meter}{a photo of a\\black car and a\\green parking meter}
    }
    \caption{\textbf{Qualitative comparison with extrinsic reward on GenEval.} Each row shows SDXL and FKS, where the final sample is the best of $K{=}4$ particles under the CLIP reward (FKS-CLIP) or under our reward (FKS-\name{}).}
    \label{fig:fks_main}
    \end{minipage}%
    \hfill
    \begin{minipage}[t]{0.55\linewidth}
    \vspace{0pt}\centering
    \includegraphics[width=\linewidth]{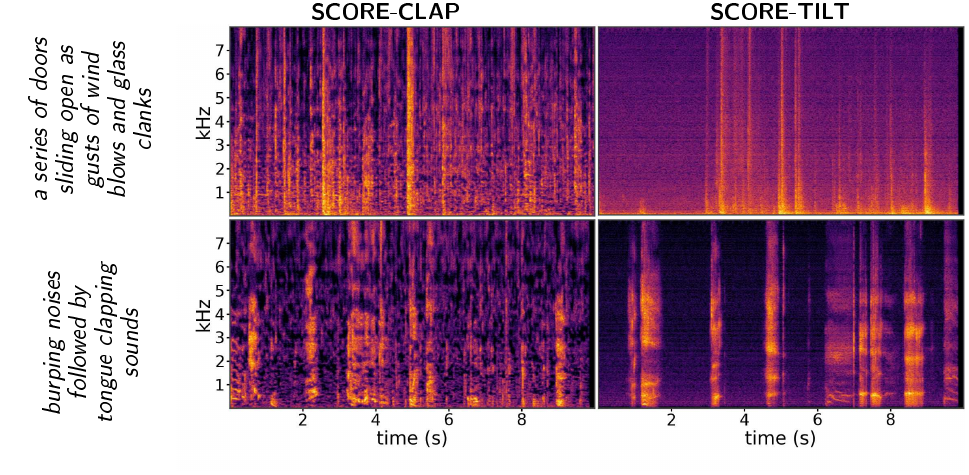}
    \caption{\textbf{Qualitative comparison with an extrinsic reward on AudioCaps.} Mel spectrograms of samples from SCORE steered by the CLAP reward (SCORE-CLAP) and by our reward (SCORE-\name{}), both on AudioLDM; each row's prompt is shown on the left. }
    \label{fig:spec_compare_main}
    \end{minipage}
\end{figure}

$\blacksquare$ \textbf{Text-to-Audio Generation.}
Because \name{} is defined entirely through the base model's own conditional distributions, it makes no assumption about the data modality. We therefore apply it unchanged to text-to-audio generation on AudioLDM~\citep{audioldm} and compare against SCORE~\citep{score}, which steers $8$ particles with an external CLAP reward, scoring them at intermediate diffusion steps and mutating them towards higher-reward samples. As shown in Table~\ref{tab:quant_audio}, AudioLDM-\name{} improves the base model on every metric (rows 1--2). Within the SCORE framework, replacing CLAP with \name{} (SCORE-\name{}) remains competitive with SCORE-CLAP on KL and FD. The comparison again favors the extrinsic reward as CLAP is trained on external audio-text pairs. Finally, to match SCORE's budget of $8$ particles, we evaluate AudioLDM-\name{} with best-of-$8$, which achieves a lower FAD than SCORE-CLAP and comparable scores elsewhere. Figure~\ref{fig:spec_compare_main} compares spectrograms on two prompts and shows how SCORE-\name{} captures the sound events better than SCORE-CLAP. More visualizations in Appendix~\ref{app:audio_qualitative}.

\begin{table}[t]
\small
\centering
\caption{Quantitative results on audio generation using prompts from AudioCaps.}
\vspace{-0.1in}
\label{tab:quant_audio}
\begin{tabular}{lcccc}
\toprule
Method & CLAP $\uparrow$ & FAD $\downarrow$ & KL $\downarrow$ & FD $\downarrow$ \\
\midrule
AudioLDM & 0.4508 & 4.9867 & 0.7782 & 3.0924 \\
AudioLDM-\name{} & 0.5350 & 3.6415 & 0.6213 & 2.7377 \\
AudioLDM-\name{} (best of 8) & \seccolor 0.6110 & \firstcolor 3.0848 & 0.6176 & 1.9624 \\
\midrule
SCORE-\name{} (8 particles) & 0.5205 & 3.8210 & \seccolor 0.5890 & \seccolor 1.8640 \\
SCORE-CLAP (8 particles) & \firstcolor 0.6258 & \seccolor 3.1874 & \firstcolor 0.5852 & \firstcolor 1.7321 \\
\bottomrule
\end{tabular}
\vspace{-0.2in}
\end{table}

\section{Related Work}
\label{sec:related}
\vspace{-0.15in}
$\blacksquare$ \textbf{Trajectory-based compositional generation.}
A growing body of work improves multi-concept generation by modifying pretrained diffusion models at inference time. Composable Diffusion~\citep{compose-diff} combines conditional scores, with MCMC-based methods improving sampling from the composed distribution~\citep{rrr}. SuperDiff~\citep{superdiff} enables principled products and mixtures, while R2F~\citep{r2f}, TweedieMix~\citep{tweediemix}, and DOS~\citep{dos} use prompt- or trajectory-level interventions. More recently, CO3~\citep{co3} suppresses modes shared by joint and single-concept distributions using a concept-contrasting corrector. 
In contrast, we define the desired pure-mode target directly in data space and sample from it via reward steering.
$\blacksquare$ \textbf{Attention and layout-based compositional control.}
Another line of work manipulates internal attention or imposes explicit
spatial structure. Attention-based approaches modify token-level activations
to reduce missing objects and attribute misbinding~\citep{ae,structured-diffusion,attention-refocus},
while layout-conditioned methods add bounding boxes, masks, region prompts,
or conditioning modules for spatial
control~\citep{box-diff,dense-diffusion,loco-layout,instance-diffusion,
gligen,t2iadapter,controlnet}. Our method requires neither attention
manipulation nor explicit layouts.
$\blacksquare$ \textbf{Reward steering and model-intrinsic rewards.}
Inference-time reward guidance steers pretrained diffusion models without retraining. DITTO~\citep{ditto} differentiates through sampling, Feynman--Kac methods use particle-based steering~\citep{fksteering,fkc}, and SCORE~\citep{score} combines multiple rewards, typically externally specified. Model-intrinsic approaches instead derive rewards from the generator itself: SAIL~\citep{sail} uses self-generated preferences, while SRUM~\citep{srum} and Self-SpectraReward~\citep{selfspectrareward} use the model to evaluate generated samples. In contrast, we derive rewards directly from conditional likelihoods of a fixed pretrained diffusion model for training-free test-time steering.

\section{Limitations and Future Work}
\vspace{-0.15in}
In our current formulation 
evaluating the reward and its gradient requires additional passes through the diffusion model, which makes our sampler more computationally expensive than standard guidance. We believe much of this overhead is not fundamental: cheaper and more accurate approximations of the reward should reduce it substantially, and we view this as a natural next step. 
Future work could also explore applications beyond images and audio, such as molecules and proteins which are also compositional in nature. 
More broadly, we hope this work points to a promising direction: steering generative models with rewards drawn from the models themselves. Such rewards need no external data, annotations, or modality-specific reward models, so the same principle can serve images, audio, and potentially other modalities alike. We are optimistic that this line of research will lead to reward steering methods that are both data-efficient and modality-agnostic.

\newpage
\subsection*{Reproducibility Statement}
All method implementations and inference pipelines are available via our project website, including instructions for environment setup, dependencies, and reproducible random seeds. The datasets used in our experiments are publicly available. We
provide detailed descriptions of the dataset in Sec. 4. Hyperparameters, inference schedules, and evaluation
pipelines are described in Sec. 4, with further details in the Appendix~\ref{app:implementation}.

\subsection*{AI Use Statement}

Generative AI tools were used to polish manuscript writing.

\bibliography{iclr2027_conference}
\bibliographystyle{iclr2027_conference}

\clearpage
\appendix
\section{Appendix}
\subsection*{Contents}
\startcontents[appendix]
\printcontents[appendix]{}{2}{\setcounter{tocdepth}{3}}

\subsection{Impact Statement}
\label{app:impact}
This paper presents work whose goal is to advance the field of text-to-image generation. There are many potential societal consequences of our work, none of which we feel must be specifically highlighted here.

\subsection{Derivations}
\label{app:deriavtions}
\subsubsection{Derivation of Proposition~\ref{prop:m2-tc}}
\label{app:tc-derivation}

\begin{restatedprop}{prop:m2-tc}{Pure-mode reward and concept synergy}
Let $w_i > 0$ with $\sum_i w_i = 1$, and let
$R_w(x_0) \coloneq \log \frac{p^\theta(x_0 \mid C)}{\prod_{i} p^\theta(x_0 \mid c_i)^{w_i}}$
be the reward in~\eqref{eq:m2-rl}, with maximizer $p^*_C$ (written $p^*$ when $C$ is fixed).
Assume the conditionals $p^\theta(x_0 \mid C)$ and $p^\theta(x_0 \mid c_i)$ are induced by a single
joint $p^\theta(x_0, c_1, \dots, c_K)$ over samples and the concept tuple $C = (c_1, \dots, c_K)$
(a \emph{consistent implicit joint}). Then:

\noindent (i) \textit{Pointwise objective.} For a fixed prompt $C$, applying Bayes' rule to $R_w$ gives,
up to a constant independent of $x_0$,
\begin{equation}
R_w(x_0) \;=\;
\underbrace{\log \frac{p^\theta(C \mid x_0)}{\prod_{i} p^\theta(c_i \mid x_0)}}_{\text{concept synergy (pcTC) } \iota_C(x_0)}
\;+\;
\underbrace{\sum_{i} (1 - w_i) \log p^\theta(c_i \mid x_0)}_{\text{per-concept evidence } \pi_w(x_0)}
\;+\; \mathrm{const}(C),
\label{eq:app-m2-tc}
\end{equation}
where $\iota_C$ is the \emph{pointwise conditional total correlation} (pcTC).
Hence solving~\eqref{eq:m2-rl} amounts to maximizing the expected pcTC plus per-concept evidence
while staying close to $p^\theta(\cdot \mid C)$, and the solution $p^*$ attains at least the
pretrained model's value of this quantity:
\begin{equation}
\E_{x_0 \sim p^*}\big[\iota_C(x_0) + \pi_w(x_0)\big] \;\ge\; \E_{x_0 \sim p^\theta(\cdot \mid C)}\big[\iota_C(x_0) + \pi_w(x_0)\big].
\label{eq:app-m2-tc-exp}
\end{equation}

\noindent (ii) \textit{Population level.} Let prompts be drawn from the prompt marginal $p(C)$ of the
consistent implicit joint, with concepts taking values in a finite vocabulary and the weights
$w$ shared across prompts. At the population level (across prompts and samples),
\begin{equation}
\E_{C \sim p(C)}\,\E_{x_0 \sim p^\theta(\cdot \mid C)}\big[R_w(x_0)\big] \;=\;
\underbrace{\mathrm{TC}(c_1; \dots; c_K \mid X_0)}_{\text{conditional Total Correlation}}
\;-\;
\underbrace{\sum_{i} (1 - w_i)\, H(c_i \mid X_0)}_{\text{per-concept uncertainty}}
\;+\; \mathrm{const},
\label{eq:app-m2-tc-pop}
\end{equation}
where $\mathrm{TC}(c_1; \dots; c_K \mid X_0) \coloneq \sum_{i} H(c_i \mid X_0) - H(c_1, \dots, c_K \mid X_0)$
is the conditional Total Correlation~\citep{watanabe_tc}, with entropies taken under the joint
$p(C)\,p^\theta(x_0 \mid C)$ read in the recognition direction: given a sample, which prompts explain it?
\end{restatedprop}

\noindent\underline{\textbf{Proof of (i).}} Write $q \coloneq p^\theta(\cdot \mid C)$.

Since both conditionals are induced by the consistent implicit joint, Bayes' rule gives
\begin{align}
\log p^\theta(x_0 \mid C)   &= \log p^\theta(C \mid x_0)   + \log p^\theta(x_0) - \log p^\theta(C), \nonumber\\
\log p^\theta(x_0 \mid c_i) &= \log p^\theta(c_i \mid x_0) + \log p^\theta(x_0) - \log p^\theta(c_i).
\label{eq:app-bayes}
\end{align}
Substituting into $R_w$, the unconditional term appears with coefficient $1 - \sum_i w_i = 0$ and
drops out:
\begin{equation}
R_w(x_0) = \log p^\theta(C \mid x_0) - \sum_i w_i \log p^\theta(c_i \mid x_0) + \kappa_w(C),
\qquad
\kappa_w(C) \coloneq \sum_i w_i \log p^\theta(c_i) - \log p^\theta(C).
\end{equation}
Adding and subtracting $\sum_i \log p^\theta(c_i \mid x_0)$ and regrouping,
\begin{align}
R_w(x_0) - \kappa_w(C)
&= \Big[\log p^\theta(C \mid x_0) - \sum_i \log p^\theta(c_i \mid x_0)\Big]
 + \Big[\sum_i \log p^\theta(c_i \mid x_0) - \sum_i w_i \log p^\theta(c_i \mid x_0)\Big] \nonumber\\
&= \underbrace{\log \frac{p^\theta(C \mid x_0)}{\prod_{i} p^\theta(c_i \mid x_0)}}_{\text{concept synergy (pcTC) } \iota_C(x_0)}
\;+\;
\underbrace{\sum_{i} (1 - w_i) \log p^\theta(c_i \mid x_0)}_{\text{per-concept evidence } \pi_w(x_0)},
\label{eq:app-decomp}
\end{align}
which is~\eqref{eq:app-m2-tc} with $\mathrm{const}(C) = \kappa_w(C)$.

$\iota_C(x_0)$ is the per-realization analogue of conditional Total Correlation, in the same way
pointwise mutual information is the per-realization analogue of mutual information~\citep{watanabe_tc}.
For a fixed prompt, it measures how much more likely the model's posterior at $x_0$ finds the concepts
\emph{jointly} than it would if they were explained independently.
Equivalently, since $\sum_i w_i = 1$,
\begin{equation}
\iota_C(x_0) + \pi_w(x_0) \;=\; \sum_i w_i \log \frac{p^\theta(C \mid x_0)}{p^\theta(c_i \mid x_0)},
\end{equation}
which is strongly negative exactly when some single concept $c_i$ explains $x_0$ much better than
the composed prompt $C$. These are the overlap modes of Fig.~\ref{fig:concept_dominance}.

Because~\eqref{eq:app-decomp} holds pointwise, for every $p$,
\begin{equation}
\E_{x_0 \sim p}[R_w(x_0)] \;=\; \E_{x_0 \sim p}\big[\iota_C(x_0) + \pi_w(x_0)\big] + \kappa_w(C).
\end{equation}
The two objectives differ by a constant, so maximizing either over the KL ball in~\eqref{eq:m2-rl}
gives the same maximizer $p^*$. The base model $q$ lies in the ball, since
$\mathrm{KL}(q \,\Vert\, q) = 0$. Hence $\E_{p^*}[R_w] \ge \E_q[R_w]$, and subtracting $\kappa_w(C)$
gives~\eqref{eq:app-m2-tc-exp}. Finally, substituting~\eqref{eq:app-decomp} into the closed
form~\eqref{eq:m2-pstar}, $\kappa_w(C)$ is absorbed into the normalizer:
\begin{equation}
p^*(x_0) \;\propto\; q(x_0)\, \exp\!\big(\beta\,[\iota_C(x_0) + \pi_w(x_0)]\big).
\end{equation}
So $p^*$ reweights the base conditional exponentially toward samples with high pcTC and high
per-concept evidence, and away from the overlap modes. \hfill$\square$

\noindent\underline{\textbf{Proof of (ii).}}

We average~\eqref{eq:app-decomp} over prompts $C \sim p(C)$ and samples $x_0 \sim p^\theta(\cdot \mid C)$.
Since $p(C)\,p^\theta(x_0 \mid C) = p^\theta(x_0)\,p^\theta(C \mid x_0)$, this is an average over samples
and the prompts that explain them (the recognition direction), so each log-posterior averages to a
negative conditional entropy:
\begin{equation}
\E\big[\log p^\theta(C \mid x_0)\big] = -H(c_1, \dots, c_K \mid X_0),
\qquad
\E\big[\log p^\theta(c_i \mid x_0)\big] = -H(c_i \mid X_0).
\end{equation}
Substituting into $\iota_C$ and $\pi_w$ and adding back the constant,
\begin{equation}
\E_{C}\,\E_{x_0 \sim p^\theta(\cdot \mid C)}\big[R_w(x_0)\big]
\;=\;
\underbrace{\sum_i H(c_i \mid X_0) - H(c_1, \dots, c_K \mid X_0)}_{\mathrm{TC}(c_1; \dots; c_K \mid X_0)}
\;-\;
\sum_{i} (1 - w_i)\, H(c_i \mid X_0)
\;+\; \mathrm{const},
\label{eq:app-tc-pop}
\end{equation}
which is~\eqref{eq:app-m2-tc-pop}, with $\mathrm{const} = \E_C[\kappa_w(C)]$. \hfill$\square$

\subsubsection{Derivation of the simplified reward}
\label{app:reward-derivation}
We derive~\eqref{eq:energy_reward}, writing $\hat{x}_0$ for the Tweedie estimate $\hat{x}_0(x_t)$ and
$R$ for the weighted reward $R_w$ of Proposition~\ref{prop:m2-tc}.

Since $\sum_i w_i = 1$, we can write $\log p^\theta(\hat{x}_0 \mid C) = \sum_i w_i \log p^\theta(\hat{x}_0 \mid C)$, so
\begin{equation}
R_w(\hat{x}_0) \;=\; \log \frac{p^\theta(\hat{x}_0 \mid C)}{\prod_i p^\theta(\hat{x}_0 \mid c_i)^{w_i}}
\;=\; \sum_{i} w_i \Big[\log p^\theta(\hat{x}_0 \mid C) - \log p^\theta(\hat{x}_0 \mid c_i)\Big].
\label{eq:app-reward-ratio}
\end{equation}
Each term compares how well the full prompt $C$ and the single concept $c_i$ explain $\hat{x}_0$.

As in diffusion classifiers~\citep{diff-classifier}, we approximate the conditional log-likelihood by the
diffusion ELBO~\citep{ddpm}, written as an expectation over timesteps $\tau \sim \mathcal{U}[0, T]$ and
noise $\eps \sim \mathcal{N}(0, I)$:
\begin{equation}
-\log p^\theta(\hat{x}_0 \mid c) \;\approx\; \E_{\tau, \eps}\Big[\gamma(\tau)\, \norm{\eps - \eps_\theta(x_\tau, \tau, c)}^2\Big] + \mathrm{const}
\;=\; -E_c(\hat{x}_0) + \mathrm{const},
\end{equation}
where $x_\tau = \sqrt{\bar{\alpha}_\tau}\, \hat{x}_0 + \sqrt{1 - \bar{\alpha}_\tau}\, \eps$ re-noises $\hat{x}_0$,
$\gamma(\tau)$ is the ELBO's time weighting,
$E_c(\hat{x}_0) \coloneq -\E_{\tau, \eps}\big[\gamma(\tau)\, \norm{\eps - \eps_\theta(x_\tau, \tau, c)}^2\big]$
is the diffusion classifier score, and the constant does not depend on the condition $c$.

Substituting into~\eqref{eq:app-reward-ratio}, the constants cancel in each log-ratio:
\begin{equation}
R_w(\hat{x}_0) \;\approx\; \sum_{i} w_i \big(E_C(\hat{x}_0) - E_{c_i}(\hat{x}_0)\big),
\end{equation}
which is~\eqref{eq:energy_reward}; any positive prefactor only rescales the reward and is absorbed into
the reward strength $\beta$ in~\eqref{eq:m2-approx-guidance}. \hfill$\square$

\noindent\textbf{Intuition.}
$E_c(\hat{x}_0)$ is high when the model denoises $\hat{x}_0$ well when told $c$. The reward is therefore
large when the full prompt $C$ explains $\hat{x}_0$ better than each single concept $c_i$ does, and small
in overlap modes, where some $c_i$ alone explains $\hat{x}_0$ as well as $C$. By Bayes' rule,
$E_C - E_{c_i} \approx \log \frac{p^\theta(C \mid \hat{x}_0)}{p^\theta(c_i \mid \hat{x}_0)} + \mathrm{const}$,
so the simplified reward is the objective of Proposition~\ref{prop:m2-tc}, computed from denoising errors.

\subsection{More implementation details}
\label{app:implementation}
For the results in Table~\ref{tab:quant_t2icomp_full} we run reward correction for the first $5$ time steps where we use \name{}-S only at the first timestep followed by \name{}-C on the next $4$ steps. At each of these timesteps, we run $5$ steps of optimization with the exception of $10$ steps at the initial timestep. We use $\beta=0.2$. 

For implementation of the energy loss, we use an adaptive weighting strategy. We use $w_i = softmax(\E_{\tau, \eps}|| \epsilon - \epsilon^\theta(c_i, \tau)||), \ i = 1, \ldots, K$ where we sample $4$ pairs of $(\eps, \tau)$ and use for all concepts $c_i$. This is the reduced variance version for the Diffusion Classifier score described in \citep{diff-classifier}. 

We use Stanza~\citep{stanza} to parse the prompts. We parse the prompts to extract different noun chunks and filter each of them to remove articles and adjectives. The remaining proper noun is used as concept in {\name}. For example, if $C$ is "a black cat and a brown dog", we consider $c_1=$"cat" and $c_2=$"dog".  \\

$\blacksquare$ \textbf{Vision Baselines} For implementation of the baselines we use their official code given in CO3~\citep{dutta2026steer}, R2F~\citep{r2f}, FKS~\citep{fksteering}, DOS~\citep{dos}, SuperDiff~\citep{superdiff}. For FKS-\name{} experiments, we used \name{} reward as a drop in replacement for the standard CLIP reward. \\

$\blacksquare$ \textbf{Audio Baselines} For audio experiments, we use the code provided in AudioLDM~\citep{audioldm} for the base model. SCORE is reimplemented from the details given in the paper~\citep{score}. For all audio experiments, we use sampling rate of $16$ kHz, DDIM with $50$ steps, CFG scale of $3.5$.

\subsection{Additional Vision Qualitative results}
\label{app:vis_qualitiative}

\begin{figure}[ht]
    \centering
    \newlength{\scImgW}\newlength{\scLabW}\newlength{\scGap}%
    \setlength{\scGap}{1pt}%
    \setlength{\scLabW}{36pt}%
    \setlength{\scImgW}{\dimexpr(\linewidth-\scLabW-7\scGap-0.5pt)/7\relax}%
    \newcommand{\scFont}{\fontencoding{T1}\sffamily}%
    \newcommand{\scVC}[1]{$\vcenter{\hbox{#1}}$}%
    \newcommand{\scHead}[1]{\hspace{\scGap}\makebox[\scImgW]{\scFont\fontseries{sb}\fontsize{7}{8}\selectfont #1}}%
    \newcommand{\scPrompt}[1]{\makebox[\scLabW]{\scVC{\rotatebox{90}{\parbox{\scImgW}{\centering\scFont\itshape\fontsize{6.5}{7}\selectfont #1}}}}}%
    \newcommand{\scImg}[1]{\hspace{\scGap}\scVC{\includegraphics[width=\scImgW]{figs/success_grid/#1.jpg}}}%
    \newcommand{\scRow}[2]{\scPrompt{#2}\scImg{#1_sdxl}\scImg{#1_sdxlcfgpp}\scImg{#1_co3}\scImg{#1_superdiff}\scImg{#1_dos}\scImg{#1_r2f}\scImg{#1_ours}\par}%
    {\offinterlineskip
    \makebox[\scLabW]{}\scHead{SDXL}\scHead{SDXL + CFG++}\scHead{CO3}\scHead{SuperDiff}\scHead{DOS}\scHead{R2F}\scHead{\fontseries{b}\selectfont TILT (Ours)}\par
    \vspace{2pt}
    \scRow{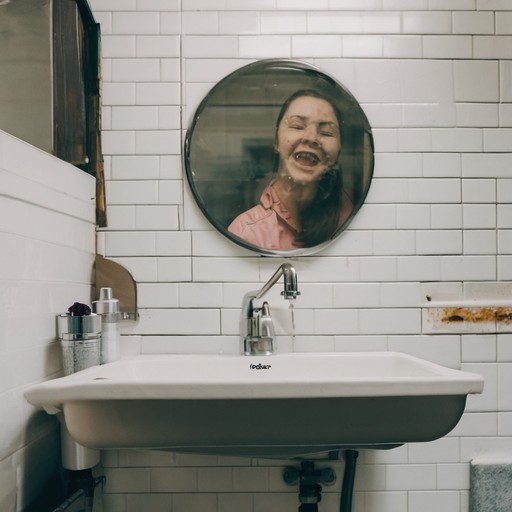}{a photo of a\\person and a sink}\vspace{\scGap}
    \scRow{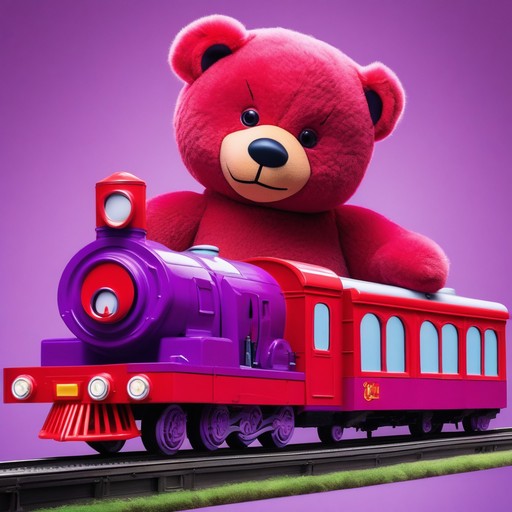}{a photo of a\\red train and a\\purple bear}\vspace{3pt}
    \scRow{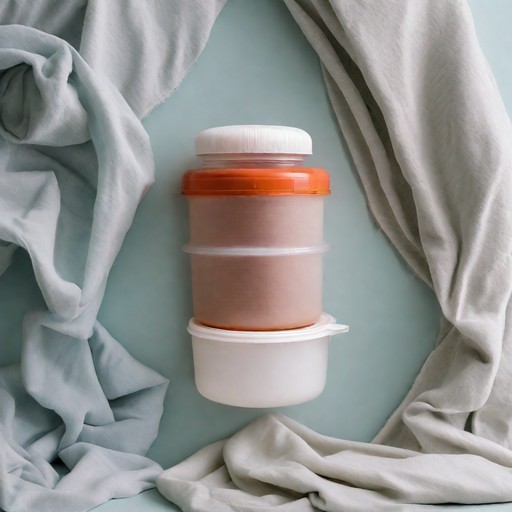}{a plastic container\\and a fabric\\blanket}\vspace{\scGap}
    \scRow{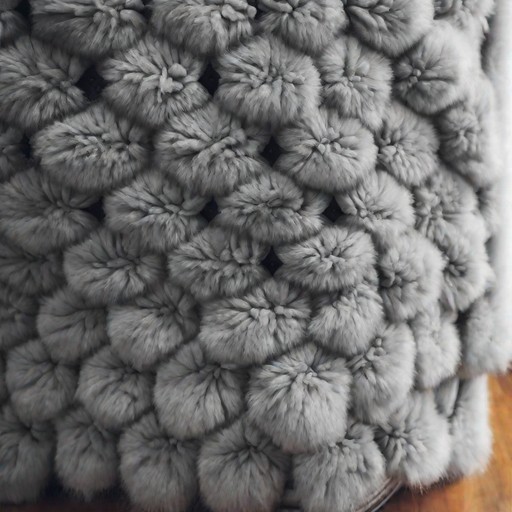}{a rubber tire\\and a fluffy rug}\vspace{\scGap}
    \scRow{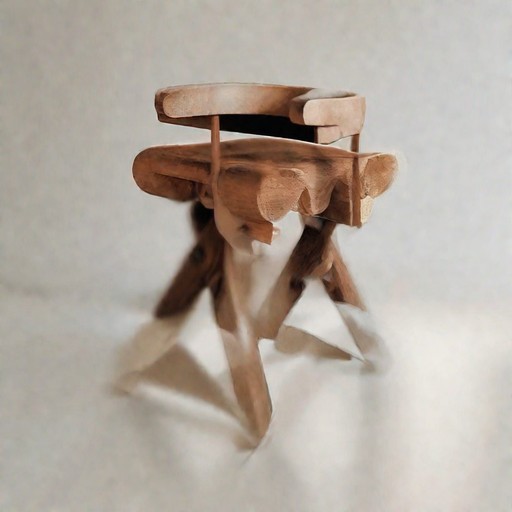}{a wooden chair\\and a fabric hat}\vspace{\scGap}
    \scRow{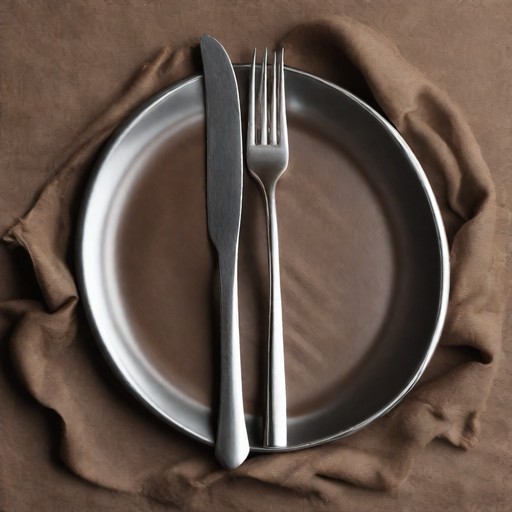}{The silver fork\\was resting on the\\brown napkin.}\vspace{\scGap}
    \scRow{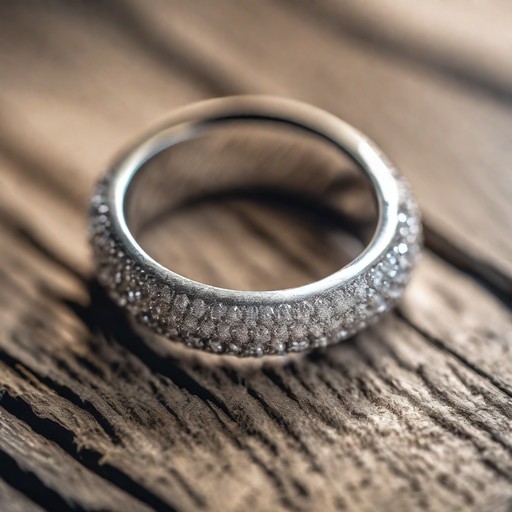}{The shiny silver\\ring sparkled on\\the rough\\wooden table.}\vspace{\scGap}
    \scRow{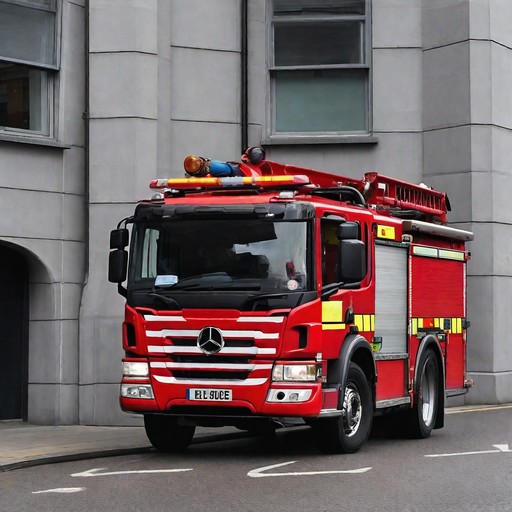}{The bright red\\fire engine sped\\past the dull\\grey building.}\vspace{\scGap}
    \scRow{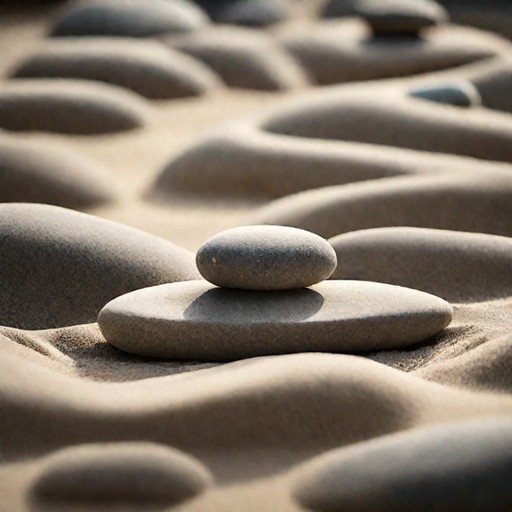}{\fontsize{5}{5.5}\selectfont The smooth, rounded stones of the zen garden lay in perfect harmony with the raked sand, inviting meditation and tranquility.}
    }
    \caption{\textbf{Additional success cases of \name{}.} The top two rows use prompts from GenEval (two objects and color attribute) and the bottom seven rows use prompts from T2I-CompBench (texture and complex subsets); each row's prompt is shown on the left. All methods use the SDXL backbone and share the same random seed within each row.}
    \label{fig:success}
\end{figure}

Figure~\ref{fig:success} shows additional prompts on which \name{} succeeds, using prompts not shown in Figure~\ref{fig:qualitative}. \name{} renders every concept with its intended attribute, while the baselines often drop one concept (e.g., the person in ``a photo of a person and a sink'') or bind the attribute to the wrong object (e.g., the purple train in ``a photo of a red train and a purple bear''). 

\subsubsection{More visual results for external reward}
\label{app:vis_success_FKS}
Figure~\ref{fig:fks_success} gives more prompts where \name{} reward has a better performance compared to the external CLIP reward, validating our previous statement of \name{}.

\begin{figure}[t] 
    \centering
    \ifdefined\fksImgW\else\newlength{\fksImgW}\newlength{\fksLabW}\newlength{\fksGap}\newlength{\fksTotW}\fi%
    \setlength{\fksGap}{1pt}%
    \setlength{\fksLabW}{28pt}%
    \setlength{\fksTotW}{0.75\linewidth}%
    \setlength{\fksImgW}{\dimexpr(\fksTotW-\fksLabW-3\fksGap-0.5pt)/3\relax}%
    \newcommand{\fksFont}{\fontencoding{T1}\sffamily}%
    \newcommand{\fksVC}[1]{$\vcenter{\hbox{#1}}$}%
    \newcommand{\fksHead}[1]{\hspace{\fksGap}\makebox[\fksImgW]{\fksFont\fontseries{sb}\fontsize{7}{8}\selectfont #1}}%
    \newcommand{\fksPrompt}[1]{\makebox[\fksLabW]{\fksVC{\rotatebox{90}{\parbox{\fksImgW}{\centering\fksFont\itshape\fontsize{6.5}{7}\selectfont #1}}}}}%
    \newcommand{\fksImg}[1]{\hspace{\fksGap}\fksVC{\includegraphics[width=\fksImgW]{#1.jpg}}}%
    \newcommand{\fksRow}[3]{\fksPrompt{#3}\fksImg{#1/cfg/#2}\fksImg{#1/fks_clip_bestof4/#2}\fksImg{#1/fks_co3r_bestof4/#2}\par}%
    {\offinterlineskip
    \makebox[\fksLabW]{}\fksHead{SDXL}\fksHead{FKS-CLIP}\fksHead{\fontseries{b}\selectfont FKS-\name{}}\par
    \vspace{2pt}
    \fksRow{figs/qualitative/fks}{02_s0_a_photo_of_a_person_and_an_apple}{a photo of a\\person and\\an apple}\vspace{\fksGap}
    \fksRow{figs/qualitative/fks}{03_s0_a_photo_of_a_fork_and_a_knife}{a photo of a\\fork and a knife}\vspace{\fksGap}
    \fksRow{figs/qualitative/fks}{04_s0_a_photo_of_a_yellow_handbag_and_a_blue_refrigerator}{a photo of a\\yellow handbag and\\a blue refrigerator}\vspace{\fksGap}
    \fksRow{figs/qualitative/fks}{05_s0_a_photo_of_a_fork_and_a_baseball_glove}{a photo of a\\fork and a\\baseball glove}\vspace{\fksGap}
    \fksRow{figs/qualitative/fks}{06_s0_a_photo_of_a_green_couch_and_an_orange_umbrella}{a photo of a\\green couch and an\\orange umbrella}
    }
    \caption{\textbf{Additional success cases of FKS with our reward on GenEval.} Each row's prompt is shown on the left. With the CLIP reward, FKS keeps samples that miss or merge an object (e.g., the person with an apple for a head, or the spoon in place of the knife) or bind a color to the wrong object (e.g., the green umbrella), while our reward keeps samples that render every object with its intended color.}
    \label{fig:fks_success}
\end{figure}

\begin{figure}[ht]
    \centering
    \newlength{\failImgW}\newlength{\failLabW}\newlength{\failGap}%
    \setlength{\failGap}{1pt}%
    \setlength{\failLabW}{28pt}%
    \setlength{\failImgW}{\dimexpr(\linewidth-\failLabW-6\failGap)/6\relax}%
    \newcommand{\failFont}{\fontencoding{T1}\sffamily}%
    \newcommand{\failVC}[1]{$\vcenter{\hbox{#1}}$}%
    \newcommand{\failHead}[1]{\hspace{\failGap}\makebox[\failImgW]{\failFont\fontseries{sb}\fontsize{7}{8}\selectfont #1}}%
    \newcommand{\failPrompt}[1]{\makebox[\failLabW]{\failVC{\rotatebox{90}{\parbox{\failImgW}{\centering\failFont\itshape\fontsize{6.5}{7}\selectfont #1}}}}}%
    \newcommand{\failImg}[1]{\hspace{\failGap}\failVC{\includegraphics[width=\failImgW]{figs/failure_grid/#1.jpg}}}%
    \newcommand{\failRow}[2]{\failPrompt{#2}\failImg{#1_sdxl}\failImg{#1_sdxlcfgpp}\failImg{#1_co3}\failImg{#1_dos}\failImg{#1_r2f}\failImg{#1_ours}\par}%
    {\offinterlineskip
    \makebox[\failLabW]{}\failHead{SDXL}\failHead{SDXL + CFG++}\failHead{CO3}\failHead{DOS}\failHead{R2F}\failHead{\fontseries{b}\selectfont TILT (Ours)}\par
    \vspace{2pt}
    \failRow{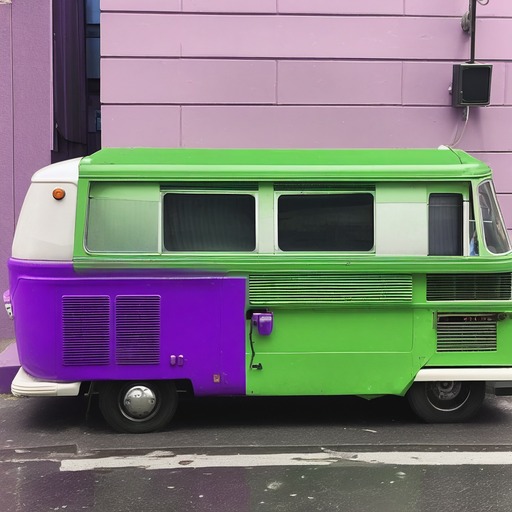}{a photo of a\\green bus and a\\purple microwave}\vspace{\failGap}
    \failRow{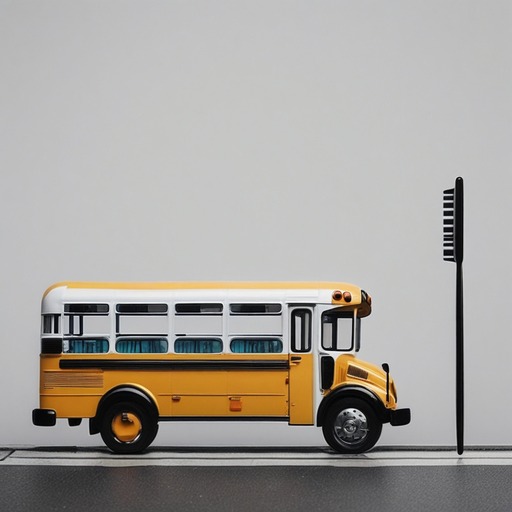}{a photo of a bus\\below a toothbrush}\vspace{\failGap}
    \failRow{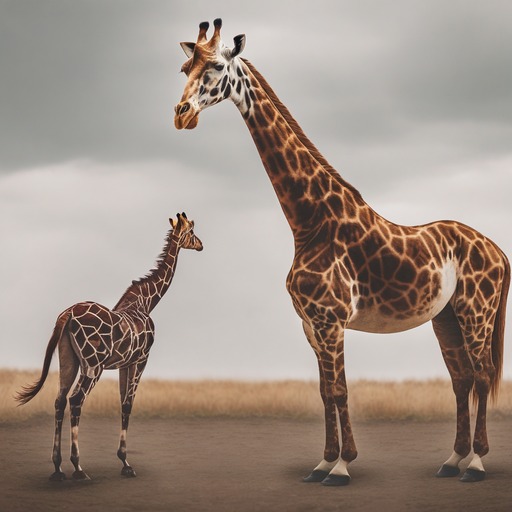}{a photo of a horse\\and a giraffe}\vspace{3pt}
    \failRow{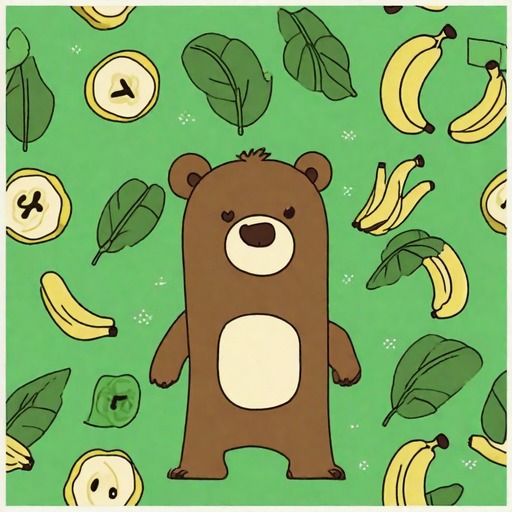}{a brown banana\\and a green bear}\vspace{\failGap}
    \failRow{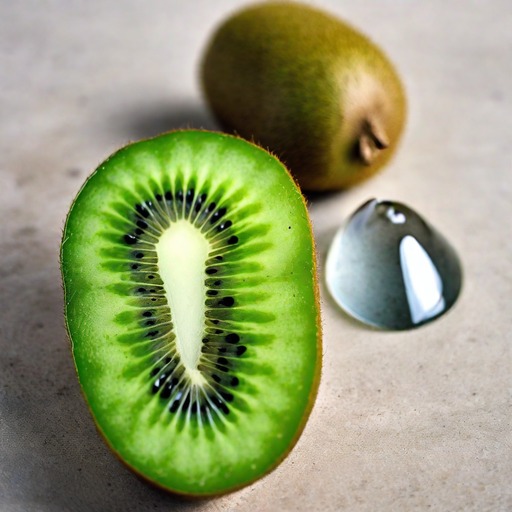}{an oblong cucumber\\and a teardrop kiwi}\vspace{\failGap}
    \failRow{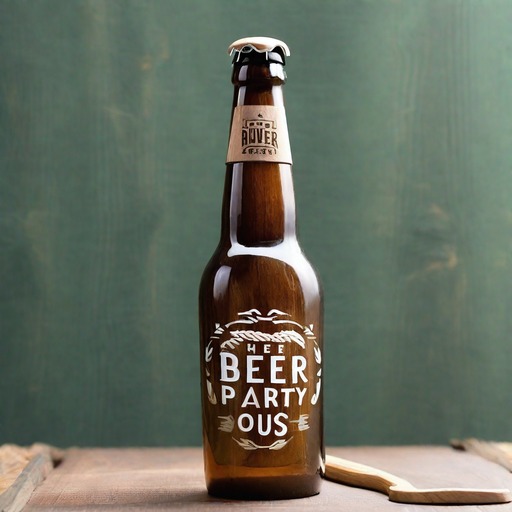}{The glass bottle and\\metallic opener hold\\the wooden beer for\\the fluffy party.}\vspace{\failGap}
    \failRow{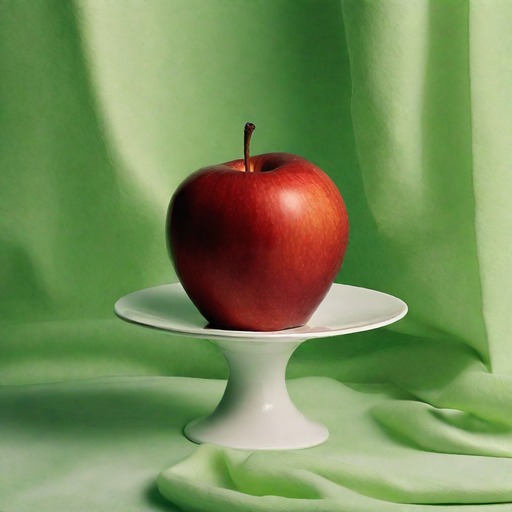}{The sharp red apple\\contrasted with\\the soft green pear.}
    }
    \caption{\textbf{Failure cases of \name{}.} The top three rows use prompts from GenEval (color attribute, position, and two objects) and the bottom four rows use prompts from T2I-CompBench (color, shape, texture, and complex subsets); each row's prompt is shown on the left. \name{} fails on these prompts for all three evaluation seeds. Typical failure modes include replacing one concept with a copy of the other, dropping a concept entirely, fusing two concepts into a hybrid and swapping attributes. }
    \label{fig:failure}
\end{figure}
\begin{figure}[t]
    \centering
    \ifdefined\fksImgW\else\newlength{\fksImgW}\newlength{\fksLabW}\newlength{\fksGap}\newlength{\fksTotW}\fi%
    \setlength{\fksGap}{1pt}%
    \setlength{\fksLabW}{28pt}%
    \setlength{\fksTotW}{0.75\linewidth}%
    \setlength{\fksImgW}{\dimexpr(\fksTotW-\fksLabW-3\fksGap-0.5pt)/3\relax}%
    \newcommand{\fksFont}{\fontencoding{T1}\sffamily}%
    \newcommand{\fksVC}[1]{$\vcenter{\hbox{#1}}$}%
    \newcommand{\fksHead}[1]{\hspace{\fksGap}\makebox[\fksImgW]{\fksFont\fontseries{sb}\fontsize{7}{8}\selectfont #1}}%
    \newcommand{\fksPrompt}[1]{\makebox[\fksLabW]{\fksVC{\rotatebox{90}{\parbox{\fksImgW}{\centering\fksFont\itshape\fontsize{6.5}{7}\selectfont #1}}}}}%
    \newcommand{\fksImg}[1]{\hspace{\fksGap}\fksVC{\includegraphics[width=\fksImgW]{#1.jpg}}}%
    \newcommand{\fksRow}[3]{\fksPrompt{#3}\fksImg{#1/cfg/#2}\fksImg{#1/fks_clip_bestof4/#2}\fksImg{#1/fks_co3r_bestof4/#2}\par}%
    {\offinterlineskip
    \makebox[\fksLabW]{}\fksHead{SDXL}\fksHead{FKS-CLIP)}\fksHead{\fontseries{b}\selectfont FKS-\name{}}\par
    \vspace{2pt}
    \fksRow{figs/failure/fks}{02_s1_a_photo_of_a_brown_dining_table_and_a_white_suitcase}{a photo of a\\brown dining table\\and a white suitcase}\vspace{\fksGap}
    \fksRow{figs/failure/fks}{04_s0_a_photo_of_a_purple_wine_glass_and_a_black_apple}{a photo of a\\purple wine glass\\and a black apple}\vspace{\fksGap}
    \fksRow{figs/failure/fks}{07_s1_a_photo_of_a_red_cake_and_a_purple_chair}{a photo of a\\red cake and a\\purple chair}\vspace{\fksGap}
    \fksRow{figs/failure/fks}{03_s1_a_photo_of_a_stop_sign_above_a_parking_meter}{a photo of a\\stop sign above\\a parking meter}\vspace{\fksGap}
    \fksRow{figs/failure/fks}{08_s1_a_photo_of_a_cell_phone_right_of_a_chair}{a photo of a\\cell phone right\\of a chair}\vspace{\fksGap}
    \fksRow{figs/failure/fks}{09_s0_a_photo_of_a_tie_right_of_a_baseball_bat}{a photo of a\\tie right of a\\baseball bat}
    }
    \caption{\textbf{Failure cases of FKS (\name{}) on GenEval.} Each row's prompt is shown on the left. FKS with our reward fails on these prompts for both evaluation seeds. The top three rows are color-attribute prompts, where the selected sample binds a color to the wrong object (e.g., the swapped colors of the cake and the chair); the bottom three rows are position prompts, where the selected sample drops an object (the parking meter) or places the objects in the wrong relation (the cell phone on the chair and the tie on the bat).}
    \label{fig:fks_failure}
\end{figure}
\begin{figure}[ht]
    \centering
    \includegraphics[width=0.9\linewidth]{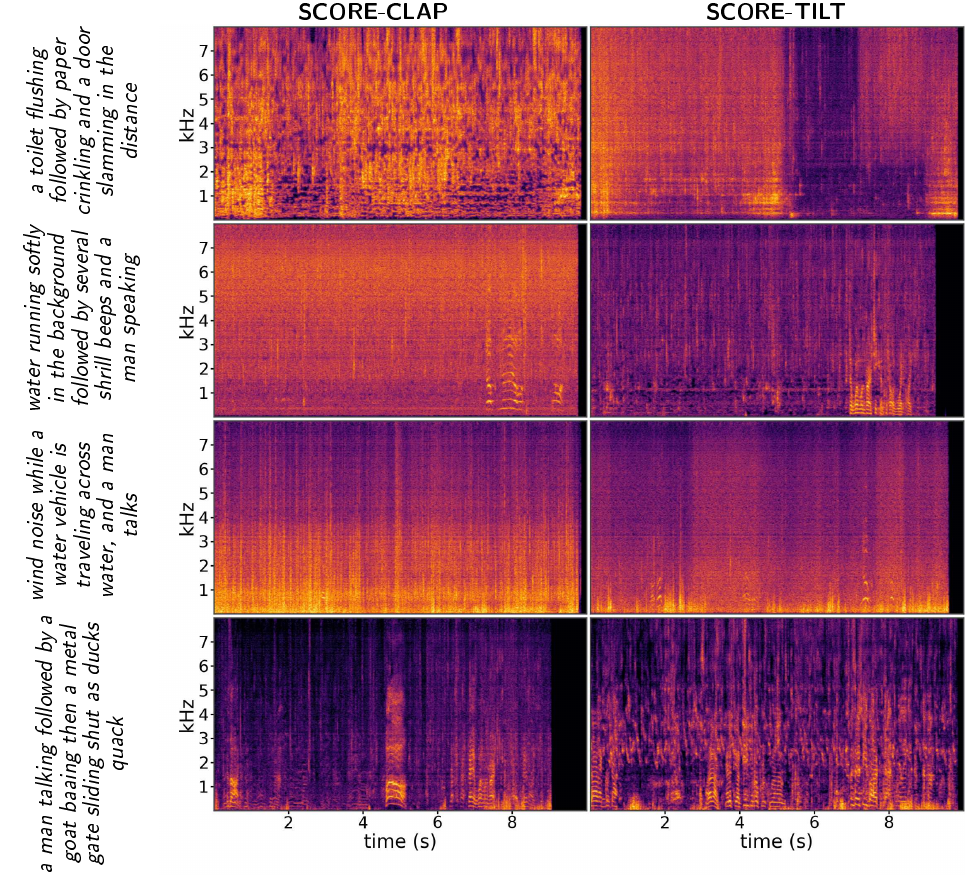}
    \caption{\textbf{Additional qualitative comparison on AudioCaps.} Mel spectrograms of samples from SCORE~\citep{score} steered by the CLAP reward (SCORE-CLAP) and by our reward (SCORE-\name{}), both on AudioLDM; each row's prompt is shown on the left.}
    \label{app_vis_audio}
\end{figure}

\subsubsection{Failure cases for \name{}}
We also present our failure cases in Fig.\ref{fig:failure} together with other baselines. Three modes of failures recur.\\
\textbf{Concept Copying.} One concept is replaced by another instance. For "a photo of a green bus and a purple microwave". 
\name{} generates a green bus and a purple bus.\\ 
\textbf{Concept Dropping.} One concept is missing completely. For prompt " a photo of a bus below a toothbrush", the toothbrush is missing.\\
\textbf{Fusion and attribute swapping.} For "a photo of a horse and a giraffe", the largest animal has a horse's body with a giraffe's coat, and the other animals are giraffes. For "a brown banana and a green bear", both objects appear but their colors are swapped.\

\subsubsection{Failure cases for comparison with external reward}
\textbf{Failure cases.} In Figure~\ref{fig:fks_failure}, we present some mismatching attribute binding and wrong spatial relation.
\subsection{Audio Qualitative results}
\label{app:audio_qualitative}
We add more qualitative results for the audio generation in Figure~\ref{app_vis_audio}.

\subsection{Ablations \& Additional Visual Numerical Results}
\label{app:numerical visual results}
Tables~\ref{tab:sv-energy-num-samples}--\ref{tab:sv-num-latent-corrector-steps} change one variable of our implementation at a time on GenEval.

\begin{table}[t]
\centering
\caption{Effect of \texttt{energy\_num\_samples}, the number of $(\tau,\epsilon)$ draws used to estimate the energy reward, on GenEval. As the number of energy samples increases, the variance of the estimate decreases, leading to more stable performance.}
\label{tab:sv-energy-num-samples}
\small
\begin{tabular}{lccc}
\toprule
\texttt{energy\_num\_samples} & BLIP-VQA $\uparrow$ & T2V $\uparrow$ & HPSv3 $\uparrow$ \\
\midrule
\multicolumn{4}{l}{\textit{Base:} default} \\
\quad 2 & 0.3942 & 0.5192 & 6.122 \\
\quad 4 & 0.3900 {\scriptsize($-$0.0041)} & 0.4952 {\scriptsize($-$0.0240$^\dagger$)} & 6.049 {\scriptsize($-$0.072)} \\
\midrule
\multicolumn{4}{l}{\textit{Base:} default with \texttt{energy\_adaptive\_weights}, \texttt{energy\_adaptive\_zeta=3}} \\
\quad 2 & 0.3898 & 0.5178 & 6.120 \\
\quad 4 & 0.3890 {\scriptsize($-$0.0009)} & 0.5067 {\scriptsize($-$0.0111)} & 6.069 {\scriptsize($-$0.051)} \\
\midrule
\multicolumn{4}{l}{\textit{Base:} default with \texttt{energy\_adaptive\_weights}, \texttt{energy\_adaptive\_zeta=5}} \\
\quad 2 & 0.3932 & 0.5194 & 6.104 \\
\quad 4 & 0.3911 {\scriptsize($-$0.0021)} & 0.5023 {\scriptsize($-$0.0171)} & 6.082 {\scriptsize($-$0.022)} \\
\midrule
\multicolumn{4}{l}{\textit{Base:} default with \texttt{energy\_adaptive\_weights}, \texttt{energy\_adaptive\_zeta=7}} \\
\quad 2 & 0.3890 & 0.5212 & 6.134 \\
\quad 4 & 0.3899 {\scriptsize($+$0.0009)} & 0.5098 {\scriptsize($-$0.0114)} & 6.093 {\scriptsize($-$0.042)} \\
\midrule
\multicolumn{4}{l}{\textit{Base:} default with \texttt{eta=0.1}, \texttt{nts\_to\_init\_correct=0}} \\
\quad 2 & 0.3945 & 0.5110 & 6.137 \\
\quad 8 & 0.3897 {\scriptsize($-$0.0047)} & 0.5062 {\scriptsize($-$0.0048)} & 6.103 {\scriptsize($-$0.034)} \\
\midrule
\multicolumn{4}{l}{\textit{Base:} default with \texttt{guidance\_scale=0.8}, \texttt{eta=0.2}, \texttt{nts\_to\_init\_correct=1},} \\
\multicolumn{4}{l}{\qquad \texttt{num\_latent\_corrector\_steps=2}, \texttt{x0\_hat\_score\_source=multi}$^\ddagger$} \\
\quad 2 & 0.3766 & 0.5093 & 6.005 \\
\quad 4 & 0.3742 {\scriptsize($-$0.0024)} & 0.4959 {\scriptsize($-$0.0134)} & 6.022 {\scriptsize($+$0.018)} \\
\bottomrule
\end{tabular}
\end{table}

\begin{table}[t]
\centering
\caption{Effect of \texttt{energy\_adaptive\_weights} defined in \secref{app:implementation}, that replaces the uniform per-concept weights of the energy reward by a softmin over the normalized score-difference norms, on GenEval.}
\label{tab:sv-energy-adaptive-weights}
\small
\begin{tabular}{lccc}
\toprule
\texttt{energy\_adaptive\_zeta} & BLIP-VQA $\uparrow$ & T2V $\uparrow$ & HPSv3 $\uparrow$ \\
\midrule
\multicolumn{4}{l}{\textit{Base:} default} \\
\quad off & 0.3942 & 0.5192 & 6.122 \\
\quad 3 & 0.3898 {\scriptsize($-$0.0043)} & 0.5178 {\scriptsize($-$0.0014)} & 6.120 {\scriptsize($-$0.002)} \\
\quad 5 & 0.3932 {\scriptsize($-$0.0010)} & 0.5194 {\scriptsize($+$0.0002)} & 6.104 {\scriptsize($-$0.018)} \\
\quad 7 & 0.3890 {\scriptsize($-$0.0051)} & 0.5212 {\scriptsize($+$0.0020)} & 6.134 {\scriptsize($+$0.013)} \\
\midrule
\multicolumn{4}{l}{\textit{Base:} default with \texttt{energy\_num\_samples=4}} \\
\quad off & 0.3900 & 0.4952 & 6.049 \\
\quad 3 & 0.3890 {\scriptsize($-$0.0011)} & 0.5067 {\scriptsize($+$0.0114)} & 6.069 {\scriptsize($+$0.019)} \\
\quad 5 & 0.3911 {\scriptsize($+$0.0011)} & 0.5023 {\scriptsize($+$0.0070)} & 6.082 {\scriptsize($+$0.033)} \\
\quad 7 & 0.3899 {\scriptsize($-$0.0001)} & 0.5098 {\scriptsize($+$0.0146$^\dagger$)} & 6.093 {\scriptsize($+$0.043)} \\
\midrule
\multicolumn{4}{l}{\textit{Base:} default with \texttt{guidance\_scale=0.8}, \texttt{eta=0.5}, \texttt{energy\_num\_samples=4}} \\
\quad off & 0.3678 & 0.5036 & 6.010 \\
\quad 3 & 0.3724 {\scriptsize($+$0.0046)} & 0.5019 {\scriptsize($-$0.0017)} & 6.004 {\scriptsize($-$0.006)} \\
\quad 5 & 0.3710 {\scriptsize($+$0.0031)} & 0.4954 {\scriptsize($-$0.0082)} & 5.995 {\scriptsize($-$0.015)} \\
\quad 7 & 0.3687 {\scriptsize($+$0.0008)} & 0.4865 {\scriptsize($-$0.0172$^\dagger$)} & 6.006 {\scriptsize($-$0.004)} \\
\bottomrule
\end{tabular}
\end{table}

\begin{table}[t]
\centering
\caption{Effect of \texttt{x0\_hat\_score\_source}, the score used to compute $\hat{x}_0$ on GenEval.}
\label{tab:sv-x0-hat-score-source}
\small
\begin{tabular}{lccc}
\toprule
\texttt{x0\_hat\_score\_source} & BLIP-VQA $\uparrow$ & T2V $\uparrow$ & HPSv3 $\uparrow$ \\
\midrule
\multicolumn{4}{l}{\textit{Base:} default} \\
\quad $\eps^{\lambda}_t$ & 0.3942 & 0.5192 & 6.122 \\
\quad $\eps_t$ \textsuperscript{\S} & 0.3937 {\scriptsize($-$0.0004)} & 0.5180 {\scriptsize($-$0.0012)} & 6.109 {\scriptsize($-$0.013)} \\
\bottomrule
\end{tabular}
\end{table}

\begin{table}[t]
\centering
\caption{Effect of \texttt{num\_ts\_to\_correct}, the number of corrected timesteps, on GenEval }
\label{tab:sv-num-ts-to-correct}
\small
\begin{tabular}{lccc}
\toprule
\texttt{num\_ts\_to\_correct} & BLIP-VQA $\uparrow$ & T2V $\uparrow$ & HPSv3 $\uparrow$ \\
\midrule
\multicolumn{4}{l}{\textit{Base:} default} \\
\quad 5 & 0.3942 & 0.5192 & 6.122 \\
\quad 2\textsuperscript{\S} & 0.3981 {\scriptsize($+$0.0040)} & 0.5087 {\scriptsize($-$0.0106)} & 6.140 {\scriptsize($+$0.019)} \\
\midrule
\multicolumn{4}{l}{\textit{Base:} default with \texttt{eta=0.1}, \texttt{nts\_to\_init\_correct=0}} \\
\quad 5 & 0.3945 & 0.5110 & 6.137 \\
\quad 10 & 0.3956 {\scriptsize($+$0.0011)} & 0.5120 {\scriptsize($+$0.0010)} & 6.126 {\scriptsize($-$0.011)} \\
\bottomrule
\end{tabular}
\end{table}

\begin{table}[t]
\centering
\caption{Effect of \texttt{num\_latent\_corrector\_steps}, the number of latent-corrector steps per corrected timestep (the first \texttt{nts\_to\_init\_correct} timesteps use \texttt{init\_latent\_corrector\_steps} instead), on GenEval. }
\label{tab:sv-num-latent-corrector-steps}
\small
\begin{tabular}{lccc}
\toprule
\texttt{num\_latent\_corrector\_steps} & BLIP-VQA $\uparrow$ & T2V $\uparrow$ & HPSv3 $\uparrow$ \\
\midrule
\multicolumn{4}{l}{\textit{Base:} default with \texttt{guidance\_scale=0.8}, \texttt{eta=0.2}, \texttt{energy\_num\_samples=4},} \\
\multicolumn{4}{l}{\qquad \texttt{nts\_to\_init\_correct=1}, \texttt{x0\_hat\_score\_source=multi}$^\ddagger$} \\
\quad 1 & 0.3725 & 0.4932 & 5.988 \\
\quad 2 & 0.3742 {\scriptsize($+$0.0017)} & 0.4959 {\scriptsize($+$0.0027)} & 6.022 {\scriptsize($+$0.035)} \\
\bottomrule
\end{tabular}
\end{table}

\end{document}